\documentclass[letterpaper, 10 pt, journal, twoside]{IEEEtran}

\IEEEoverridecommandlockouts                              

\usepackage{hyperref}

\usepackage{multicol}
\usepackage[utf8]{inputenc} 
\usepackage[T1]{fontenc}    
\usepackage{hyperref}       
\usepackage{url}            
\usepackage{booktabs}       
\usepackage{fancyhdr}
\usepackage{amsfonts}       
\usepackage{nicefrac}       
\usepackage{microtype}      
\usepackage{epigraph}
\usepackage{graphicx}
\usepackage{amsmath}
\usepackage{array}
\usepackage[binary-units=true]{siunitx}
\usepackage{subcaption}
\usepackage{bm}
\usepackage{mathtools}
\usepackage{cuted}
\usepackage{algorithm}
\usepackage{algorithmic}
\usepackage{etoolbox}
\usepackage{balance}
\usepackage{multirow}
\usepackage[skip=10pt,font=small]{caption}

\usepackage{wrapfig}
\usepackage{tikz}

\newcommand\mypara[1]{\vspace{3pt}\noindent\textbf{#1.}}

\newcommand{\lone}{\(\mathcal{L}_1\)}
\newcommand{\lonesp}{\(\mathcal{L}_1\)\ }
\usepackage{marginnote} 
\newcommand{\rebuttal}[1]{\textcolor{black}{#1}}
\usepackage{pgfplots}
\pgfplotsset{compat=newest}
\usepgfplotslibrary{patchplots}
\usepackage{grffile}
\usepackage{amsmath}
\usepackage{interval}
\usepackage{balance}

\usepgfplotslibrary{groupplots}
\usepgfplotslibrary{dateplot}

\title{\LARGE \bf

A Learning-based Quadcopter Controller with Extreme Adaptation

}

\author{Dingqi Zhang$^1$, Antonio Loquercio$^2$, Jerry Tang$^1$, Ting-Hao Wang$^1$,\\ Jitendra Malik$^3$, and Mark W. Mueller$^1$
\thanks{The authors are with the $^1$High Performance Robotics Lab, Dept. of Mechanical Engineering, UC Berkeley, $^2$ the University of Pennsylvania (most work done while at UC Berkeley), and $^3$ Dept. of Electrical Engineering and Computer Science, University of California at Berkeley. Contact at \{dingqi, loquercio, jerrytang, wtyngh, malik, mwm\}@berkeley.edu} 
}

\fancypagestyle{firstpageheader}{
  \fancyhf{}
  \fancyhead[C]{ \large \textcolor{gray}{This paper has been accepted for publication in the IEEE Transactions on Robotics (T-RO), 2025. \textcopyright~IEEE}}
  
}

\begin{document}

\maketitle
\thispagestyle{firstpageheader}


\begin{abstract}
This paper introduces a learning-based low-level controller for quadcopters, which adaptively controls quadcopters with significant variations in mass, size, and actuator capabilities. Our approach leverages a combination of imitation learning and reinforcement learning, creating a fast-adapting and general control framework for quadcopters that eliminates the need for precise model estimation or manual tuning. The controller estimates a latent representation of the vehicle's system parameters from sensor-action history, enabling it to adapt swiftly to diverse dynamics.
Extensive evaluations in simulation demonstrate the controller's ability to generalize to unseen quadcopter parameters, with an adaptation range up to 16 times broader than the training set. In real-world tests, the controller is successfully deployed on quadcopters with mass differences of 3.7 times and propeller constants varying by more than 100 times, while also showing rapid adaptation to disturbances such as off-center payloads and motor failures. These results highlight the potential of our controller to simplify the design process and enhance the reliability of autonomous drone operations in unpredictable environments.
Video and code are at: \url{https://github.com/muellerlab/xadapt_ctrl}
\end{abstract}

\begin{figure}[ht!]
    \centering
    \includegraphics[width=\linewidth]{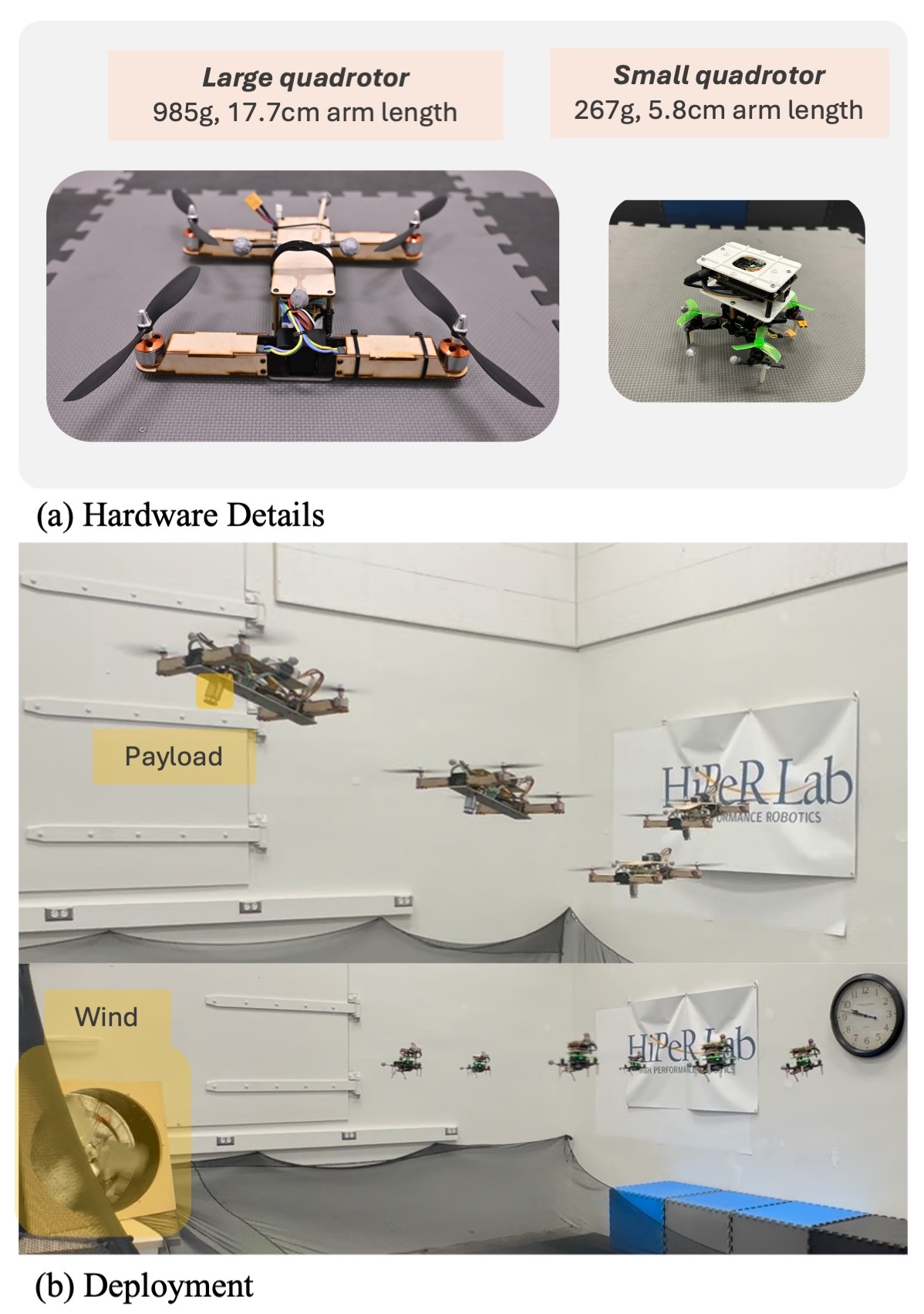}
    \caption{Our adaptive controller can control quadcopters with large difference in parameters such as mass, arm length, and actuators while also show disturbance rejection. \textbf{(a)} Two vehicles flown with our controller, with mass differing by a factor of 3.68, arm length by 3.1 and motor constant by more than 100x. \textbf{(b)} Demonstration of our controller on these two vehicles for the task of tracking trajectories under disturbances including off-center payload and wind. We use a single control policy across different drones and tasks, which is deployed without any vehicle-specific modifications.}
    \label{fig:grandFig1}
\end{figure}

\section{Introduction}

The agile nature of quadcopters and the necessity for precise control in dynamic environments create a unique context for exploring control strategies. Model-based controllers for quadcopters generally rely on estimates of the vehicle's properties, including inertia, motor constants, and other parameters. 
Notable examples include sliding mode control~\cite{ZHENG2014slidingmode} and PID controllers~\cite{mellinger2012trajectory}.
Once these parameters are estimated, the controller typically requires iterative tuning through successive experiments to refine its performance. However, inaccuracies in parameter estimation can directly lead to execution errors in controller commands. 
Furthermore, any modification to the vehicle, such as attaching an extra payload, could lead to suboptimal performance without repeating the estimation and tuning processes. Such significant engineering effort could be eased with a universal controller that does not require specialized tuning.

In this work, we propose a learning-based low-level controller designed to control a variety of quadcopters with notable differences in mass, size, propellers, and motors. Our controller is also capable of rapidly adapting to unknown in-flight disturbances such as off-center payloads, loss of efficiency in motors, and wind.

\subsection{Related Work}

The development of fixed-parameter controllers, while foundational, is inherently limited by their lack of real-time adaptivity to model uncertainties and disturbances. Adaptive control techniques were introduced to address these unpredicted variations in a system. One of the initial contributions in this field was the model reference adaptive controller (MRAC), an extension of the well-known MIT-rule~\cite{MAREELSMit}. The empirical success of MRAC led to the development of $\mathcal{L}_1$ control~\cite{cao2008design,hovakimyan2010l1}, which offers a promising solution by estimating the differences between the nominal state transitions predicted by the reference model and those observed in practice. Such differences are compensated by allocating a control authority proportional to the disturbance, effectively driving the system back to its reference behavior.

However, the performance of the classic $\mathcal{L}_1$ formulation degrades when the observed transitions deviate greatly from the (usually linear) reference model. This limitation is critical in scenarios involving large variations between different quadcopters' dynamics, as the underlying assumptions on locally linear disturbances are generally not fulfilled. \rebuttal{Some studies have discussed practical considerations regarding its implementation. For instance, high adaptive gains in certain scenarios may introduce numerical challenges and sensitivity to unmodeled dynamics~\cite{ioannou2014l1critique}. It has also been noted that, under specific conditions, the closed-loop behavior of $\mathcal{L}_1$ controllers can resemble that of linear proportional-integral (PI) controllers~\cite{ortega2014l1critique, ortega2016l1adaptunnecessary}, which may limit the expression of its adaptive features.} To address these issues, recent research has explored combining $\mathcal{L}_1$ adaptive control with nonlinear online optimization techniques, such as model predictive control (MPC)~\cite{tao2023l1mpc,hanover2021performance, pravitra2020}. These methods have achieved impressive results, but still require explicit and accurate knowledge of the reference model for adaptation.

\rebuttal{Beyond $\mathcal{L}_1$, advanced nonlinear adaptive controllers have been developed to enable agile quadcopter maneuvers in the presence of uncertainty.  Notable approaches include geometric adaptive tracking control on SE(3)~\cite{goodarzi2015geometric} and adaptive incremental nonlinear dynamic inversion (INDI)~\cite{smeur2015adaptINDI}. These controllers improve performance by directly incorporating nonlinear dynamics and real-time uncertainty estimation into the control design. A nominal model prior is first estimated and then used for subsequent compensation of dynamic variations, enhancing robustness during high-precision maneuvers.}

Recent advances in data-driven control strategies have shown promising results for quadcopter stabilization~\cite{hwangbo2017control,koch2019reinforcement}, or waypoint tracking flight~\cite{song2021autonomous, kaufmann2022benchmark}, as well as agile racing against human pilots~\cite{kaufman2023racingdrone}. Model-free reinforcement learning in~\cite{kaufman2023racingdrone} has demonstrated impressive adaptability to unmodeled disturbances, such as blade flapping effects. Combining data-driven methods with model-based control designs has also been proposed to leverage the guaranteed adaptivity and robustness offered by model-based control. For instance, some studies have learned policies from model-based methods like Robust-Tube MPC through imitation learning~\cite{tagliabue2021demonstrationefficient, zhao2023efficient,tagliabue2024efficientdeeplearningrobust}. Another approach is augmenting the learned controller with classical adaptive control designs during deployment for fast disturbance estimation and online adaptation~\cite{neuralfly,huang2023datt}. Despite these advancements, these methods remain tailored to specific platforms. Transferring the same controller to another vehicle typically requires retraining or fine-tuning the policy, along with data collection for the new vehicle.

Zero-shot adaptation across different vehicles has been demonstrated on quadrupeds~\cite{feng2022genloco}, highlighting the versatility of learning-based methods. However, this generalized control relies on existing internal motor control loops rather than directly adapting at the motor level. In the case of quadcopters, the variation in actuators between different vehicles is particularly significant, with motor constants potentially differing by orders of magnitude. Consequently, effective adaptation across different quadcopter platforms must address motor-level differences directly. Additionally, the high-frequency nature of motor control increases the risk of crashes due to inadequate adaptation. These factors, the substantial variation in actuators and the high-frequency nature of motor control, pose significant challenges to applying previous methods of adaptive trajectory control to the problem of extreme adaptation across quadcopters.

\subsection{Our Contributions}

In this work, we present a general framework for learning low-level adaptive controllers that are effective across a wide range of quadcopters. Similar to prior research in learning-based control for aerial robots~\cite{kaufmann2020RSS, kaufmann2022benchmark, eschmann2024learning, molchanov2019sim,hwangbo2017control, becker2020learning}, we train the policy entirely in simulation using reinforcement learning and deploy it directly to the real world without fine-tuning (i.e., zero-shot deployment).

While previous works typically rely on slight parameter randomizations (up to 30\%) around a nominal model~\cite{kaufmann2022benchmark}, our approach must handle parameter variations thousands of times larger. This presents a significant challenge for reinforcement learning, as such a broad range of variations can hinder optimization convergence. 

To address this challenge, we build on our earlier conference work on learning-based low-level control~\cite{zhang2023universal} and introduce three key technical innovations: (1) a dual training strategy that combines imitation learning from specialized model-based controllers and model-free reinforcement learning. This combination effectively handles the challenges of training a low-level controller due to its high-frequency nature and the low informational density of observations. (2) A specifically designed reward to provide the low-level controller with direct feedback for quick adjustments, allowing it to perform more agile maneuvers, and (3) a designed-informed domain randomization method to ensure that the variations in quadcopter designs during training are consistent with real-world constraints. 
These innovations eliminate constraints on slow flight and accurate state estimation, and widens the range of vehicles our policy can fly. Our approach significantly outperforms existing baselines.
In addition, it enables the controller to adapt to out-of-distribution quadcopters in simulation up to 16x wider than the training set and to disturbances for which it was not explicitly trained, such as wind.

Our work shows a generalized controller for agile and robust flight of quadcopters with parameter differences of several orders of magnitude.
Such large scale adaptability will help democratize the process of drone design by enabling users that lack modeling expertise to control custom-made vehicles.

\section{Method}
\label{sec:method}


We present our methodology for learning an adaptive low-level controller for various quadcopters. A list of symbols and notations are given in Table~\ref{tab:symbols}.

\begin{table}[ht!]
    \centering
    \caption{List of Symbols}
    \label{tab:symbols}
    \resizebox{\columnwidth}{!}{
    \begin{tabular}{c p{3cm} c p{3cm}}
    \toprule
         \multicolumn{2}{l}{\textbf{Notation} (for arbitrary symbol $x$)} & \multicolumn{2}{l}{} \\
        \midrule
        $ x $ & Scalar quantity & $ \bm{x} $ & Vector quantity (e.g., $ \bm{\omega} $) \\
        $ x_\mathrm{des} $ & Commanded quantity & $x_\mathrm{ref}$ & Quantity in reference trajectory \\
        $ x_{\min} $ & Minimum value & $ x_{\max} $ & Maximum value \\
        $\dot x$ & Derivative of $x$ w.r.t. time&$ \hat{x} $ & Estimated quantity\\
        \midrule
        \multicolumn{2}{l}{\textbf{State Variables}} & \multicolumn{2}{l}{} \\
        \midrule
        $\bm{p}$& Position & $\bm{v}$ & Velocity \\ $\bm{q}$ & Attitude (quaternion) &
        $\psi$ & Yaw angle \\ 
        $\bm{\omega}$ & Angular velocity & $ \bm{\tau} $ & Torque  \\
        $c_{\Sigma}$ & Mass-normalized total thrust & $\bm{F}$ & Individual motor forces\\
        $\bm{a}$ & Individual motor speeds &
        $\bm{a}_\mathrm{pwm}$ & Individual motor Pulse width modulation (PWM) commands\\ 
        \midrule
        \multicolumn{2}{l}{\textbf{Quadrotor Parameters}} & \multicolumn{2}{l}{} \\
        \midrule
        $l$ & Arm length & $m$ & Mass\\
        $ J $ & Mass moment of inertia (MMOI) matrix & 
        $C_d$ & Body drag coefficient \\
        $C_F$ &Propeller constant: thrust-to-motorspeed-squared ratio&
        $C_\tau$&Propeller constant: torque-to-thrust ratio \\
        $c$& Size factor for randomization& $\lambda$ & The sampled quadrotor 
         \\
         $M$& Mixer matrix &\\
        \midrule
        \multicolumn{2}{l}{\textbf{Learning Variables}} & \multicolumn{2}{l}{} \\
        \midrule
        $\pi$& Base policy (our low-level controller)&$\phi$& Adaptation module \\
        $\mu$ & Intrinsics encoder& $\bm{e}_t$ & Environmental parameters \\
        $\bm{z}_t$& Intrinsics vector &$\bm{x}_t$& State vector\\
        $r_t$ &Reward&&\\
        \bottomrule
    \end{tabular}%
    }
\end{table}

\subsection{Control Structure}

Cascade control systems are instrumental in managing complex dynamic systems by decomposing them into a hierarchy of simpler nested subsystems. For example, in a two-layer system, the high-level component focuses on high-level tasks such as trajectory planning, while the low-level component acts as the inner control loop to execute the commands from the high level.

Our controller is designed to function as a low-level component within this modular control hierarchy. It translates high-level total thrust and angular velocity commands into individual motor speeds with adaptation to different quadcopters. 
%
%
The implementation of low-level adaptation abstracts away the physical complexities of the system from the high-level planner, allowing the latter to focus on high-level mission tasks. This flexibility also enables our controller to enhance non-adaptive high-level controllers, adding adaptability to disturbances and model mismatches and thereby improving overall system performance.

\begin{figure}
\centering
\includegraphics[width=\linewidth]{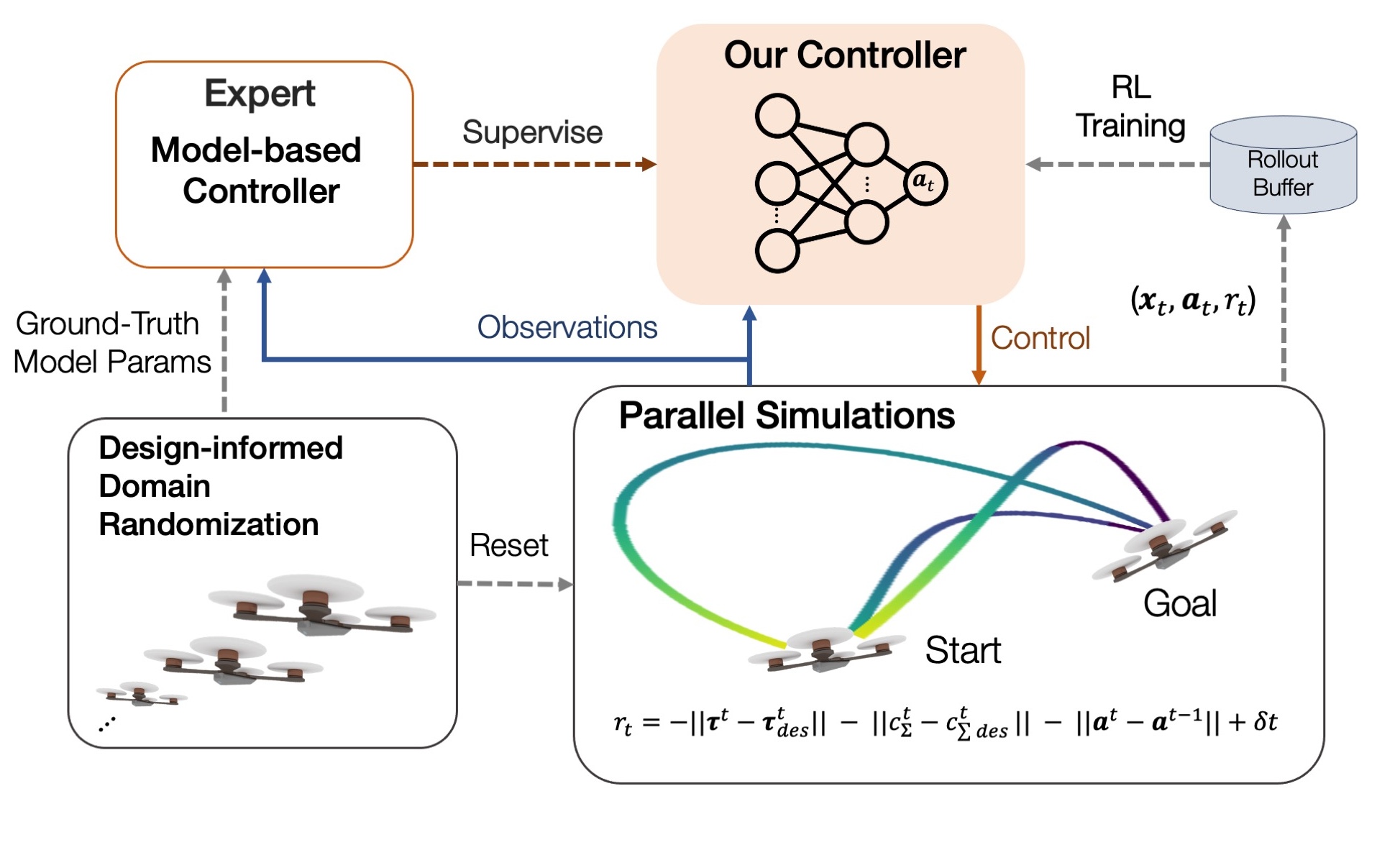}
\caption{An overview of the training process for our adaptive controller. The policy aims to track reference trajectories in simulation for various different quadcopters. The training framework employs a hybrid approach, combining reinforcement learning (RL) with imitation learning derived from a model-based controller. We use a design-informed randomization strategy to generate various quadcopters that adhere to general quadcopter design principles
}
\label{fig:high-level-method}
\end{figure}

An overview of our training strategy is illustrated in Figure~\ref{fig:high-level-method}. 
In simulation, the controller learns to track diverse trajectories across various quadcopter models. The training follows a dual approach, combining reinforcement learning (RL) with imitation learning (IL) from an expert model-based controller.
To ensure efficient and realistic training, we employ a design-informed randomization law to sample quadcopter parameters. This approach maintains adherence to real-world design constraints while enhancing training efficiency. The generated parameters are then used to update the expert model-based controller, ensuring it adapts its behavior as the simulated model changes.

The following subsections provide details on each component of the training process.

\subsection{Training Framework}

\begin{figure}[t]
\centering
\includegraphics[width=\linewidth]{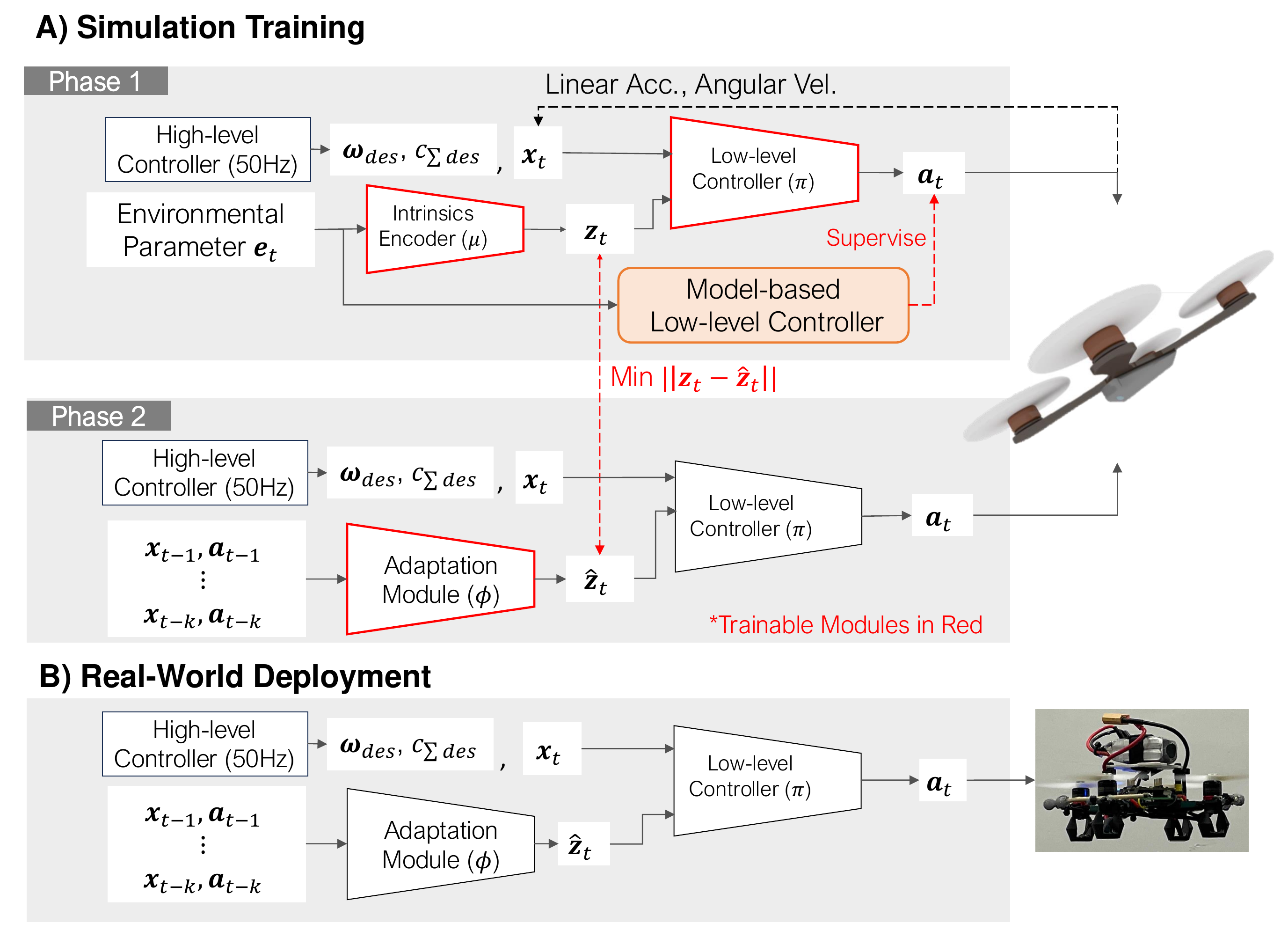}
\caption{\rebuttal{The training (top) and the deployment architecture of our system (bottom). We train in two phases. In the first phase, we train a low-level controller $\pi$ via RL and IL. The policy $\pi$ takes the current state $\bm x_t$ and the intrinsic vector $\bm z_t$, which is a compressed version of the environment parameters $\bm e_t$ generated by the module $\mu$. Since we cannot deploy this policy in the real world because the environment parameters $e_t$ are not available, we learn an adaptation module that takes the sensor-action history and directly predicts the intrinsics vector $\bm z_t$. This is done in phase two in simulation using supervised learning. We can finally deploy the base policy $\pi$ which takes as input the current state $\bm x_t$ and the intrinsics vector $\hat{\bm z_t}$ predicted by the adaptation module $\phi$. }}
\vspace{-2ex}
\label{fig:method}
\end{figure}

\rebuttal{We adopt a learning-based framework that decouples policy learning into a control policy and a real-time estimator, commonly used in quadruped locomotion~\cite{kumar2021rma, lee2020learning}. 
The detailed training process is shown in Figure~\ref{fig:method}. We train in two phases. In the first, we train a low-level controller $\pi$ given access to ground-truth system parameters via RL and IL. Since we don't have such parameters in the real world, we use a second phase to learn an adaptation module $\phi$. Such a module predicts the system parameter from a sensor-action history. The module is trained in simulation using supervised learning. During deployment, we can use the base policy $\pi$ and the adaptation module $\phi$ to achieve zero-shot adaptation.}

%
\mypara{Phase 1 Training} Our controller consists of a base policy $\pi$, an intrinsics encoder $\mu$, and an adaptation module $\phi$. At time $t$, the base policy $\pi$ takes the current state $\bm x_t \in \mathbb{R}^{8}$ and the ground-truth intrinsics vector $\bm z_t  \in \mathbb{R}^8$ to output the target motor speeds $\bm a_t \in \mathbb{R}^4$ for all individual motors.
The intrinsics vector $\bm{z}_t$ is a low dimensional encoding of the environment parameters $\bm{e}_t \in \mathbb{R}^{34}$, which consist of model parameters or external disturbances that are key to adaptive control. 
We use the intrinsics encoder $\mu$ to compress $\bm e_t $ to $\bm z_t$. This gives us:
\begin{align}
        \bm z_t &= \mu(\bm e_t) \\  
        \bm a_t &= \pi(\bm x_t, \bm z_t)
\end{align}
The current state $\bm{x}_t$ includes the mass-normalized thrust $c_\Sigma$ ($\mathbb{R}$), angular velocity $\bm \omega$ ($\mathbb{R}^3$), commanded total thrust $c_{\Sigma,\mathrm{des}}$ ($\mathbb{R}$) and commanded angular velocity $\bm{\omega}_{\mathrm{des}}$ ($\mathbb{R}^3$).
The environmental parameter $\bm e_t$ includes mass $m$, arm length $l$, propeller constants (torque-to-thrust ratio $C_\tau$ and thrust-to-motorspeed-squared ratio $C_F$), the diagonal entries of MMOI matrix $J$ ($\mathbb{R}^3$), body drag coefficients $C_d$ ($\mathbb{R}^{3}$), maximum motor rotation, motor effective factors ($\mathbb{R}^{4}$), mixer matrix $M$ ($\mathbb{R}^{4\times4}$), payload mass, and external torque ($\mathbb{R}^{3}$), which results in an 34 dimensional vector. \rebuttal{The mixer matrix $M$ is the inverse of the allocation matrix. While the allocation matrix maps rotor forces to total thrust and torques, the mixer matrix maps desired thrust and torques to individual rotor forces, making it essential for motor control to determine the commands for individual rotors. A more detailed derivation is provided, for example, in~\cite{mahony2012dronemodeling}.}

The choice of 8 dimensions for the intrinsics vector $\bm{z}_t$ is determined empirically through a binary search on the dimension of the intrinsics $\bm{e}_t$, aiming to optimize the learning performance of the policy. 
The latent representation of high-dimensional system parameters allows the base policy to adapt to variations in drone parameters, payloads, and disturbances such as external force or torque. 

\mypara{Phase 2 Training} During deployment, we do not have access to the environmental parameters $\bm e_t$ and hence we cannot directly measure the intrinsics $\bm{z}_t$ in the real world. Instead, we estimate it via the adaptation module $\phi$, which uses the commanded actions and the measured sensor readings from the latest $k$ steps to estimate it online during deployment as~\eqref{eq:phi}. 
We can train this adaptation module in simulation using supervised learning because we have access to the ground truth intrinsics $\bm z_t$. We minimize the mean squared error loss $\lVert \bm z - \hat{\bm z} \rVert ^2$ when $\hat{\bm z}$ is estimated using sensor-action history of the vehicle tracking trajectories generated with randomized motion primitives~\cite{mueller2015trajectory}. The random trajectory tracking task can provide a set of rich excitation signals for the adaptation module to estimate $\hat{\bm z}$ using the sensor-action history.

The estimated $\hat{\bm{z}_t}$ along with the current state $\bm{x}_t$ is fed into our base policy $\pi$ to output motor speed during deployment as~\eqref{eq:pi}. More concretely, 
\begin{align}
\hat{\bm{z}_t} &=  \phi\big(\bm{x}_{t-k:t-1}, \bm{a}_{t-k:t-1}\big) \label{eq:phi}\\
\bm{a}_t &= \pi(\bm{x}_t, \hat{\bm{z}_t}) \label{eq:pi}
\end{align}

\subsection{Reward Design}\label{sec:reward-des}

The adaptive base policy functions as a low-level controller within the control hierarchy, capable of tracking arbitrary high-level commands irrespective of the vehicle being controlled.
Our reward design should align with this objective by incentivizing the agent to track the specified reference high-level commands and penalizing crashes and oscillating motions. 

The reward at time $t$ is calculated as the sum of the following quantities:
\begin{enumerate}
    \item Output Smoothness Penalty: $-\| \bm{a}^{t} - \bm{a}^{t-1} \|$
    \item Survival Reward: $\delta t$
    \item Mass-normalized Thrust Tracking Deviation Penalty: $-\| {c}^{t}_{\Sigma} - {c}_{\Sigma,\mathrm{des}}^{t} \|$
    \item Torque Tracking Deviation Penalty: \\$-\| \bm{\tau}^{t} - \bm{\tau}_\mathrm{des}^{t} \|$
\end{enumerate}
Where $\delta t$ is the simulation step in the training episode. The mass-normalized thrust command $c_{\Sigma,\mathrm{des}}$ is given by the high-level controller along with the commanded angular velocity $\bm{\omega}_\mathrm{des}$. The commanded torque $\bm{\tau}_\mathrm{des}$ is given by the rate control on the commanded angular velocity. In particular, 
\begin{align}
    \dot{\bm \omega}_{\mathrm{des}} &= K(\bm{\omega}_\mathrm{des} - \bm{\omega}) \label{eq:des-omega-dot}\\
    \bm{\tau}_\mathrm{des} &= J\dot{\bm \omega}_{\mathrm{des}}  + \bm{\omega} \times (J \bm{\omega}) \label{eq:des-torque}
\end{align}
Where $K$ is a diagonal gain matrix, which we choose with values $K = \text{diag}(20, 20, 4)$s$^{-1}$ to effectively control the torques in roll, pitch, and yaw axes individually. The higher gains for roll and pitch (20) prioritize their control over yaw (4), due to their greater importance for flight stability and maneuverability.

The output oscillation penalty encourages smooth inputs, discouraging high-frequency control signals.
The survival reward encourages the quadcopter to learn to fly longer until the end of the training episode.
Finally, both tracking deviation penalties encourage the quadcopter to track the given high-level commands by matching its mass-normalized thrust and torque with the reference commands.

\subsection{Guiding search by imitating a model-based controller}\label{sec:base_policy}

We implement the base policy $\pi$ and the intrinsics encoder $\mu$ as multi-layer perceptrons and jointly train them end-to-end in simulation. The training is done by an integration of IL and model-free RL.
\rebuttal{In our approach, the expert controller is a low-level proportional-derivative (PD) controller with access to the sampled vehicle's ground truth model parameters. It receives mass-normalized thrust and angular velocity commands from the high-level controller and computes the desired torque using~\eqref{eq:des-torque}. The resulting thrust and torque are linearly mapped to individual motor forces ($\bm{F}_\mathrm{des}$) using the mixer matrix ($M$), which are then converted into motor speeds ($\bm{a}_\mathrm{exp}$), representing the expert action.}
\rebuttal{
\begin{align} \label{eq:thrust-mapping}
    \bm{F}_\mathrm{des} &= M \begin{bmatrix}
        m {c}_{\Sigma,\mathrm{des}} \\
        \bm{\tau}_\mathrm{des}
    \end{bmatrix} \\
    \label{eq:spd-mapping}
    \bm a_\mathrm{exp} &= \sqrt{\frac{\bm{F}_\mathrm{des}}{C_F}} 
\end{align}
where the square root operation is applied elementwise.}

The key distinction from other work combining reinforcement and imitation learning~\cite{xie2020learning,huang2024manipulator,loquercio2023learning} is that during each training episode, the ground-truth model parameters of randomized quadcopters are used to adapt the expert controller. This ensures that our base policy learns from an expert controller that dynamically adjusts its behavior whenever the quadcopter model changes. 
The IL loss minimizes the mean squared error loss on actions:
\begin{align}
    L_{IL}(\pi) = \lVert \bm a_\mathrm{exp} - \bm a \rVert ^2
\end{align}

Reinforcement learning maximizes the following expected return of the policy $\pi$: 
\begin{align}
    R_{RL}(\pi) = \mathbb{E}_{\tau \sim p(\tau|\pi)}\Bigg[\sum_{t=0}^{T-1}\gamma^t r_t\Bigg]
\end{align}
where $\tau = \{(x_0, a_0, r_0), (x_1, a_1, r_1) . . .\}$ is the trajectory of the agent when executing the policy $\pi$, and $p(\tau|\pi)$ represents the likelihood of the trajectory under $\pi$.

We adaptively change the relative weight between these two losses. The overall training framework seeks to maximize the overall reward of the policy $\pi$:
\begin{align}
        R(\pi) &=(1-\alpha) R_{RL}(\pi)-\alpha L_{IL}(\pi) \\
        \alpha &= e^{-0.001t_\text{epoch}}
\end{align}  
where the weight of the IL losses decays exponentially while the weight for RL increases inversely with training steps so that RL becomes dominant later in the training process.
This training scheme enables rapid learning of the desired behavior from the expert controller at the beginning of training and generalization by RL in the later parts of training.

\subsection{Quadcopter Parametric Randomization} \label{sec:quad-rand}
\begin{table}[t]
\caption{Ranges of quadcopter and environmental parameters, along with the end states of the sampled trajectories from the initial conditions. Parameters without units are dimensionless.}
\label{tab:randomization}
\setlength{\tabcolsep}{3pt} 
\begin{center}
\begin{tabular}{lcc}
\toprule
\textbf{Parameters} & \textbf{Training Range} & \textbf{Testing Range} \\
\midrule
\textbf{Quadcopter Parameters} & & \\
Mass (kg) & [0.226, 0.950] & [0.205, 1.841] \\
Arm length (m) & [0.046, 0.200] & [0.040, 0.220] \\
MMOI around $x$, $y$ (kg$\cdot$m$^2$) & [1.93e-4, 5.40e-3] & [1.73e-5, 2.27e-2] \\
MMOI around $z$ (kg$\cdot$m$^2$) & [2.42e-4, 8.51e-3] & [2.10e-4, 3.40e-2] \\
\begin{tabular}[c]{@{}l@{}}Propeller constant: \\Torque-to-Thrust Ratio (m)\end{tabular}
 & [0.0069, 0.0161] & [0.0051, 0.0170] \\
Payload (\% of Mass) & [18, 40] & [18, 40] \\
\begin{tabular}[c]{@{}l@{}}Payload location from\\Center of Mass\\(\% of Arm length)\end{tabular} & [-50, 50] & [-50, 50] \\
\begin{tabular}[c]{@{}l@{}}Propeller Constant:\\Thrust-to-Motorspeed-squared Ratio\end{tabular} & [3.88e-8, 8.40e-6] & [3.24e-9, 1.02e-4] \\
Body drag coefficient & [0, 0.74] & [0, 1.15] \\
Max. motor speed (rad/s) & [800, 8044] & [400, 10021] \\
Motor effectiveness factor & [0.7, 1.3] & [0.7, 1.3]\\
Motor time constant (s) & 0.01 & 0.01 \\
\hline
\multicolumn{3}{l}{\textbf{Sampled Trajectory End State from Initial Condition}} \\
Position (m) &[-2, 2]& [-2, 2] \\
Velocity (m/s) &[-2, 2]& [-2, 2]\\
Acceleration (m/s$^2$) &[-2, 2]& [-2, 2] \\
Total Time (s) & [1, 5] & 5 \\
\bottomrule
\end{tabular}
\end{center}
\end{table}

The training of our adaptive policy requires a wide spectrum of quadcopters, a challenge that we address through a carefully crafted randomization process rather than relying on uniform sampling.
We propose a randomization method that embodies key physical principles and design constraints of quadcopters, ensuring that the generated variations are physically plausible.
Quadcopters follow a general design pattern, which typically involves a symmetric structure with four rotors positioned at the corners of a square frame. 
The size of a quadcopter is positively correlated with its mass, moment of inertia, and other properties, such as motor power and body drag coefficient. 
Our method follows the pattern in randomizing the quadcopters and their respective dynamic characteristics, instead of simply varying parameters independently. 
In particular, we introduce a few key factors in quadcopter randomization which govern the variation of some other quadcopter body parameters. 

\mypara{Size Factor} 
We introduce a size factor $c$, which uniformly scales the size and motor strength of the quadcopter.   
We randomly sample $c$ from the range of [0, 1]. The arm length is linearly scaled with $c$ with minimum and maximum values from the training range of Table~\ref{tab:randomization}.
\begin{align}
    l = c(l_{\max}-l_{\min}) + l_{\min}
\end{align}
Assuming a constant density and proportional scaling in all dimensions, the mass of the quadcopter is proportional to its volume, which in turn scales with the cube of its arm length.
Similarly, the moment of inertia, which depends on both the mass distribution and the distance from the axis of rotation, scales approximately with the fifth power of the arm length under these assumptions. The body drag coefficient, primarily influenced by the cross-sectional area the quadcopter presents to the airflow, scales with the square of the arm length.

We preserve the correlation by defining the mass, moment of inertia and body drag coefficient as
\begin{align}
&\begin{cases}
    m &= c_m(m_{\max} - m_{\min}) + m_{\min}\\
    c_m &= \frac{l^3 - l_{\min}^3}{l_{\max}^3 - l_{\min}^3} 
\end{cases}\\
&\begin{cases}
        J &= c_J(J_{\max} - J_{\min}) + J_{\min}\\
    c_J &= \frac{l^5 - l_{\min}^5}{l_{\max}^5 - l_{\min}^5} 
\end{cases} \\
&\begin{cases}
C_d &= c_{C_d}(C_{d_{\max}} - C_{d_{\min}}) + C_{d_{\min}}\\
c_{C_d} &= \frac{l^2 - l_{\min}^2}{l_{\max}^2 - l_{\min}^2} 
\end{cases}
\end{align}
with all minimum and maximum values from Table~\ref{tab:randomization}.

To reflect the relationship between quadcopter size and motor strength, we choose to exponentially scale the motor thrust-to-motorspeed-squared ratio with the size factor. This design choice ensures that larger quadcopters, which typically require more powerful motors, are equipped with appropriately scaled motor capabilities in our simulations.
\begin{align}
    C_F = C_{F_{\min}} \left( \frac{C_{F_{\max}}}{C_{F_{\min}}}\right) ^c
\end{align}
Finally, all other parameters, such as maximum motor speed and propeller constant, are linearly scaled with the size factor. 
This factor and the associated randomization method ensure the correlation between quadcopter parameters, reducing the likelihood of generating physically unrealistic quadcopters (e.g., a very small and lightweight quadcopter equipped with overly powerful motors).

\mypara{Noise} To ensure flexibility and allow for the fact that real systems will not perfectly follow these scaling rules, we introduce a uniformly distributed noise in the range of [-20\%, 20\%] to all parameters after they have been scaled with the size factor $c$.

\mypara{Motor Effectiveness Factor} We randomize the motor effectiveness factor for each of the four rotors so that the motor will produce a force different than expected. For each rotor, the simulated motor speed is calculated by multiplying the intended speed by this factor.
This is to simulate the motor ineffectiveness due to battery voltage drop, a damaged propeller, or simply hardware variations.

\mypara{External Disturbance} At a randomly sampled time during each episode, the parameters, including mass, inertia, and the center of mass, are again randomized. 
This is used to mimic sudden variations in the quadcopter parameters due to a sudden disturbance caused by an off-center payload.

 All our training and testing ranges in simulation are listed in Table~\ref{tab:randomization}.

\section{Implementation Details} \label{sec:ex-setup}
This section details the specific implementation of our approach, including the simulator for the training and evaluation of the policy, the hardware specifications for real-world experiments, and the neural network architectures with their training details.

\mypara{Simulation Environment} We use the Flightmare simulator~\cite{song2020flightmare} to train and test our control policies. 
We implement the same high-level controller in \cite{mueller2018multicopter} to generate high-level commands at the level of body rates and mass-normalized collective thrust for our low-level controller to track.
It is designed as a cascaded linear acceleration controller with desired acceleration mimicking a spring-mass-damper system with natural frequency 2rad/s and damping ratio 0.7. The desired acceleration is then converted to the desired total thrust and the desired thrust direction, and the body rates are computed from this as proportional to the attitude error angle, with a time constant of 0.2s.
%
The high-level controller's inputs are the platform's state (position, rotation, angular, and linear velocities) and the reference position, velocity, and acceleration from the generated trajectory at the simulated time point. 
The policy outputs individual motor speed commands, and we model the motors' response using a first-order system with a time constant of 10ms.
%
Each RL episode lasts for a maximum of 5s of simulated time, with early termination if the vehicle loses more than 10m from its starting height, or the quadcopter's body rate exceeds 10rad/s.
The control frequency is 500Hz, which is also the simulation step.
We additionally implement an measurement latency of 5ms.

\mypara{Hardware Details} For all of our real-world experiments, we use two quadcopters, which differ in mass by a factor of 3.68, and in arm length by a factor of 3.1. 
The first one, which we name \emph{large quadrotor} has a mass of 985g, a size of 17.7cm in arm length, a thrust-to-weight ratio of 3.62, a diagonal inertia matrix of [0.004, 0.008, 0.012]kg$\cdot$m$^2$ (as expressed in the z-up body-fixed frame), and a maximum motor speed of 1000rad/s. The second one, \emph{small quadrotor}, has a mass of 267g, a size of 5.8cm in arm length, a thrust-to-weight ratio of 3.23, a diagonal inertia matrix of [259e-6, 228e-6, 285e-6]kg$\cdot$m$^2$, and a maximum motor speed of 6994rad/s.
For each of our platforms, we use a Qualcomm Robotics RB5 platform as the onboard computer which runs the high-level control at 50Hz and our deployed policy at 500Hz, and a mRo PixRacer as the flight control unit.  
%
%
We use as high-level a PID controller which takes as input the goal position, velocity, and acceleration and outputs the mass normalized collective trust and the body rates.
An onboard Inertia Measurement Unit (IMU) measures the angular velocity and the acceleration of the robot, which is low-pass filtered to reduce noise and remove outliers.
The high-level commands of the collective thrust and the body rates, and the low-level measurement of the angular rates and the acceleration are fed into the deployed policy as inputs.
The policy outputs motor speed commands, which are sent to the PixRacer via the UART serial port and subsequently tracked by off-the-shelf electronic speed controllers.
\rebuttal{The real-world experiment is performed indoors with a motion capture system running.
The motion capture is only used for evaluating the system's performance in experiments and for feedback to the high-level controller, but it does not provide any feedback to our policy.}

\mypara{Network Architecture and Training Procedure} The base policy is a 3-layer MLP with 256-dim hidden layers. This takes the drone state and the vector of intrinsics as input to produce motor speeds. The environment factor encoder is a 2-layer MLP with 128-dim hidden layers. The policy and the value function share the same factor encoding layer. The adaptation module projects the latest 100 state-action pairs into a 128-dim representation, with the state-action history initialized with zeros. \rebuttal{We selected a window size of 100 as it provides good performance while keeping the network lightweight.} Then, a 3-layer 1-D CNN convolves the representation across time to capture its temporal correlation. The input channel number, output channel number, kernel size and stride of each CNN layer are [32, 32, 8, 4], [32, 32, 5, 1], [32, 32, 5, 1]. The flattened CNN output is linearly projected to estimate the intrinsics vector $\bm z_t$. For RL, we train the base policy and the environment encoder using PPO~\cite{schulman2017proximal} for 100M steps in PyTorch. We use the reward described in Section~\ref{sec:reward-des}. Policy training takes approximately 1.5 hours on an ordinary desktop machine with 1 NVIDIA GeForce RTX 4060 GPU.
We then train the adaptation module with supervised learning by rolling out the student policy. We train with the ADAM optimizer to minimize MSE loss. 
We run the optimization process for 10M steps, training on data collected over the last 1M steps. Training the adaptation module takes approximately 20 minutes.
Both networks are trained with the deep learning framework PyTorch. For more efficient inference and resource allocation on the onboard computer, we use Mobile Neural Network (MNN)~\cite{jiang2020mnn, proc:osdi22:walle} to convert trained models to MNN formats to optimize their inference speed and overhead. \rebuttal{Table~\ref{tab:mnn_infer} presents the average inference time measured on the RB5 platform for onboard computation. To meet the required 500Hz control frequency, the combined inference time of the controller policy and adaptation module must be less than or equal to 2ms. MNN satisfies this requirement with a total average inference time of 0.165ms, whereas PyTorch exceeds the limit with a total of 78.850ms, making real-time onboard inference infeasible.}

\begin{table}[h]
\caption{\label{tab:mnn_infer}\rebuttal{The average inference time for the PyTorch and MNN frameworks measured over a 10-second window on the Qualcomm Robotics RB5 platform.}}
\centering
\rebuttal{ 
\begin{tabular*}{0.45\textwidth}{@{}lc|c@{}}
\toprule
    & \multicolumn{2}{c}{\begin{tabular}[c]{@{}c@{}}Mean Inference Time$\pm \sigma$ (ms)\end{tabular}}\\
\midrule
    & PyTorch&MNN  \\
Low-level Controller ($\pi$)  &73.626$\pm$2.692&0.078$\pm$0.003 \\
Adaptation Module ($\phi$) & 5.224$\pm$0.708&0.087$\pm$0.005 \\
\bottomrule
\end{tabular*}
}
\end{table}

\section{Simulation Experiments}
In this section, we evaluate the performance of our controller through multiple simulation experiments.
We begin by establishing a set of baseline methods and justifying their selection. Subsequently, we evaluate each method and ours on the task of trajectory tracking for randomized quadcopters. 
The results of these initial tests motivate us to further challenge our approach on quadcopters that significantly deviate from the training distribution.
The simulation experiments offer a controlled environment to assess our approach on aspects of robustness, adaptivity and generalization, thus paving the way for subsequent hardware experiments.

\subsection{Baselines Setup}
\begin{figure}[t]
    \centering
    \includegraphics[width=\linewidth]{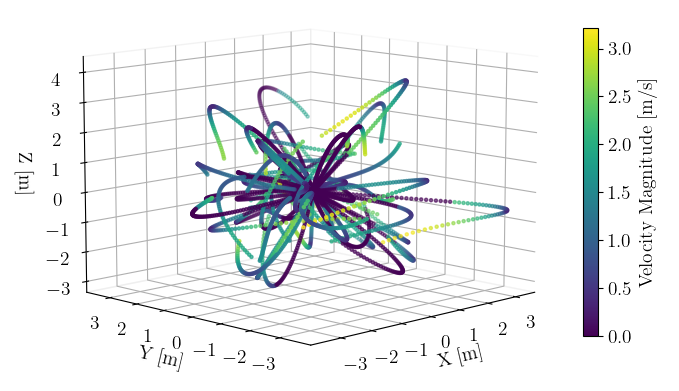}
    \caption{Visualization of typical desired quadcopter trajectories in 3D space during simulated tests. The color gradient represents the magnitude of the velocity at each point along the trajectory. We sample 50 trajectories from the origin with the distribution defined by Table~\ref{tab:randomization}. The initial conditions of all trajectories are hovering at origin.}
    \label{fig:test-trajectory-visual}
\end{figure}

We compare our approach with a set of baselines in the simulation. 
The task is to evaluate the tracking performance of a randomly sampled quadcopter along random trajectories. 
We randomize quadcopters according to our design-informed domain randomization technique outlined in Section~\ref{sec:quad-rand}. 
The testing range is listed in Table~\ref{tab:randomization} and a sample of typical desired trajectories is shown in Figure~\ref{fig:test-trajectory-visual}. 
We choose a nominal quadcopter model $\lambda_{\mathrm{norm}}$, which is obtained by setting $c=0.5$ when sampling without noise added.  

We choose several different sets of high-level and low-level controllers as baselines from prior work. \rebuttal{We include the implementation details of all baselines in Appendix~\ref{app:baseline-impl}. Most
baseline names follow the format \textit{high-level controller\textbf{-}low-
level controller}, with exceptions explicitly noted.} First, we include \textit{PID-PD$_*$} and \textit{PID-PD$_n$} as reference baselines. \textit{PID-PD$_*$} serves as an upper performance bound, as its low-level PD controller has access to the ground-truth model parameters of each sampled quadcopter. \rebuttal{This makes it equivalent to the expert controller used during training, as described in Section~\ref{sec:base_policy}.} In contrast, \textit{PID-PD$_n$} serves as a lower performance bound, since its low-level PD controller only uses nominal parameters from the model $\lambda_{\mathrm{norm}}$ and does not adapt to individual quadcopters. These two baselines provide a sanity check: any effective adaptation method should achieve performance within this range. \rebuttal{To ensure that the adaptation task is handled entirely by the low-level controller, all high-level controllers in both the baselines and our framework use the same control parameters across all tested vehicles. These control gains are tuned specifically to optimize performance on the nominal model $\lambda_{\mathrm{norm}}$.}

Next, we include \textit{\lone-PD$_n$} and \textit{PID-\lone} as additional baselines for evaluating our method. 
\textit{\lone-PD$_n$} employs the \lonesp adaptive high-level controller (\cite{cao2008design,hovakimyan2010l1}) to compensate for model uncertainties while keeping the low-level PD controller fixed.
In contrast, \textit{PID-\lone} applies \lonesp as a low-level adaptive controller. These baselines serve a dual purpose: they act as additional benchmarks for evaluating our approach while also validating our core assumption about adaptation.
We design our method as a low-level controller under the assumption that adaptation across the diverse parametric range in Table~\ref{tab:randomization} is more effective when handled at this control level. 
In other words, an adaptive high-level controller alone cannot sufficiently compensate for the model disparity in our problem, which involves controlling quadcopters with significant differences in design and actuators. The \lonesp baselines allow us to test this assumption by comparing their performance when adaptation occurs at different control levels. 

\rebuttal{Finally, we compare our method to state-of-the-art adaptive controllers. We include \textit{Geo-A}, a geometric adaptive controller that operates at both high and low levels~\cite{goodarzi2015geometric} so is an exception to our naming convention. We also include \textit{PID-INDI-A}, which combines a PID high-level controller with a low-level adaptive INDI controller~\cite{smeur2015adaptINDI}. These baselines provide further context on how our method compares to established nonlinear adaptive control techniques.}

At the beginning of each experiment, the quadcopter is spawned with a hovering state. The trajectory to track is generated with the motion primitive generation algorithm~\cite{mueller2015trajectory} with the end condition sampled from the test range in Table~\ref{tab:randomization}. The experiment is considered successful if the position tracking error is within 2m at every point of the trajectory. 

\subsection{Results} \label{sec:sim-baseline-results}

\begin{table}[h]
\caption{\label{tab:highlevel}\rebuttal{We choose 6 baselines: \textit{PID-PD$_*$}, \textit{PID-PD$_n$}, \textit{\lone-PD$_n$}, \textit{PID-\lone}, \textit{Geo-A} and \textit{PID-INDI-A}}. The \textit{PID-PD$_*$} has access to all ground-truth system parameters and thus could be regarded as the expert. We compare their performance on the task of tracking random quadcopters along trajectories. The test ranges are defined in Table~\ref{tab:randomization}. The metrics are the success rate, \rebuttal{the maximum position tracking error}, the position and velocity RMSE between the actual quadcopter trajectory and the reference trajectory. The results are from 100 experiment for each baseline. }
\centering
\begin{tabular*}{0.49\textwidth}{@{}lcccc@{}}
\toprule
    & Success & \rebuttal{Max Pos.} &Position&  Velocity\\
    &Rate&\rebuttal{Err (m)}& $\mathrm{RMSE} \pm \sigma$  (m)& $\mathrm{RMSE} \pm \sigma$(m/s) \\
\midrule
PID-PD$_n$ & 22\% & \rebuttal{1.565} & 0.510$\pm$0.372 & 0.845$\pm$1.066 \\
\lone-PD$_n$ & 62\% & \rebuttal{1.105} & 0.186$\pm$0.167 & 0.278$\pm$0.392 \\
\rebuttal{Geo-A} & \rebuttal{64\%} &\rebuttal{1.033} & \rebuttal{0.128$\pm$0.160 } & \rebuttal{0.174$\pm$0.263 }\\
\rebuttal{PID-INDI-A} & \rebuttal{67\%} &\rebuttal{0.333} & \rebuttal{\textbf{0.063$\pm$0.060} } & \rebuttal{\textbf{0.075$\pm$0.082} }\\
PID-\lone & 77\% & \rebuttal{1.304} & 0.221$\pm$0.242 & 0.357$\pm$0.481 \\
\textbf{PID-Ours} & \textbf{100\%} & \rebuttal{\textbf{0.311}} &0.148$\pm$0.075 & 0.129$\pm$0.066 \\
\midrule
\begin{tabular}[c]{@{}l@{}}
PID-PD$_*$\\
(Expert)
\end{tabular}& 100\% & \rebuttal{0.221} & 0.061$\pm$0.057 & 0.059$\pm$0.050 \\
\bottomrule
\end{tabular*}
\end{table}

The results of the simulation experiments are reported in Table~\ref{tab:highlevel}. We compare the five approaches under three metrics: (i) the success rate, \rebuttal{(ii) the maximum position tracking error}, (iii) the root-mean-square error (RMSE) in position tracking, and (iv) the RMSE in velocity tracking. We rank the methods according to the success rate and the tracking performance.

Given the very large amount of quadcopter variations, \textit{PID-PD$_n$} with only access to the nominal model $\lambda_{\mathrm{norm}}$ achieved the lowest success rate and the largest tracking error. In contrast, \textit{PID-PD$_*$} has a 100\% success rate with the lowest tracking error, since it uses the ground-truth parameters of the quadcopter in computing the control inputs. Without access to the ground truth parameters as \textit{PID-PD$_*$} but with adaptation to the unknown dynamics, the flight performance of \lonesp controllers significantly increase. However, with adaptation at high-level, the \textit{\lone-PD$_n$} achieves a lower success rate than its counterpart \textit{PID-\lone} with adaptation at low-level. Since tracking errors are only computed in successful runs, the \textit{\lone-PD$_n$} achieves a slightly lower tracking error. This result has shown that an adaptive low-level controller tends to perform better with the large model disparity across the platforms, which justifies our assumption for our controller design. 

\rebuttal{The two nonlinear adaptive baselines \textit{Geo-A} and \textit{PID-INDI-A} both show better tracking performance than \lonesp controllers. Specifically, \textit{PID-INDI-A} achieves the lowest tracking error in successful flights among all methods. However, both baselines achieve a similar success rate of approximately 65\%. Specifically, we often observe them failing near the boundaries of the adaptation range, where the dynamics deviate significantly from the nominal model. Moreover, their performance relies on prior knowledge of the reference model $\lambda_{\mathrm{norm}}$, and \textit{INDI-A} in particular also requires access to current motor speed measurements and accurate estimates of angular acceleration and torque, whereas our method operates without such information. Despite this, our method achieves a 100\% success rate and the lowest maximum position error among all baselines, with only a slightly higher average tracking error than the expert controller with access to the true parameters. } 

\subsection{Generalization} \label{sec:generalization}

\begin{figure}[t]
    \centering
    \includegraphics[width=0.8\linewidth]{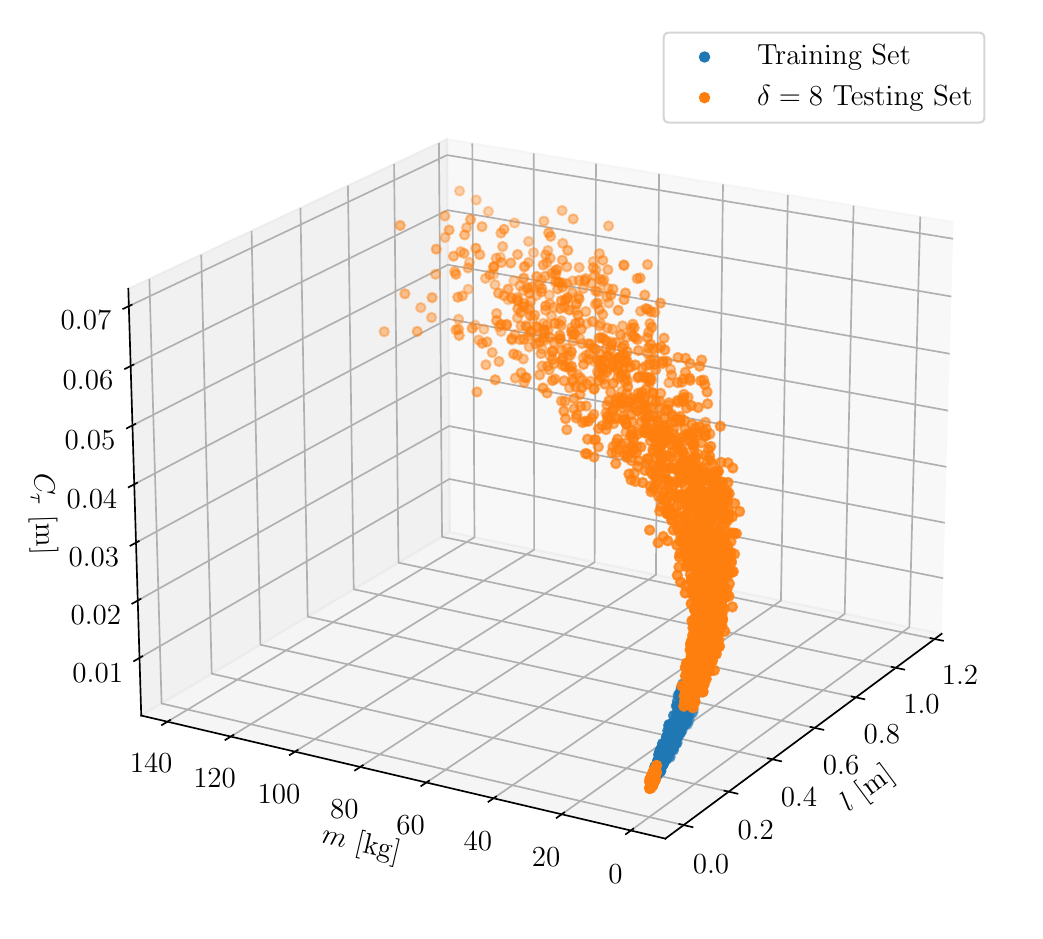}
    \caption{Visualization of the differences between the training set and the $\delta=8$ testing set. The scatter plot shows 2,000 randomly sampled quadcopters based on arm length $l$, mass $m$, and torque-to-thrust coefficient $C_{\tau}$. The testing set exhibits a significantly wider distribution than the training set.}
    \label{fig:train-test-visual}
\end{figure}

\begin{figure}[t]
\centering
\includegraphics[width=\linewidth]{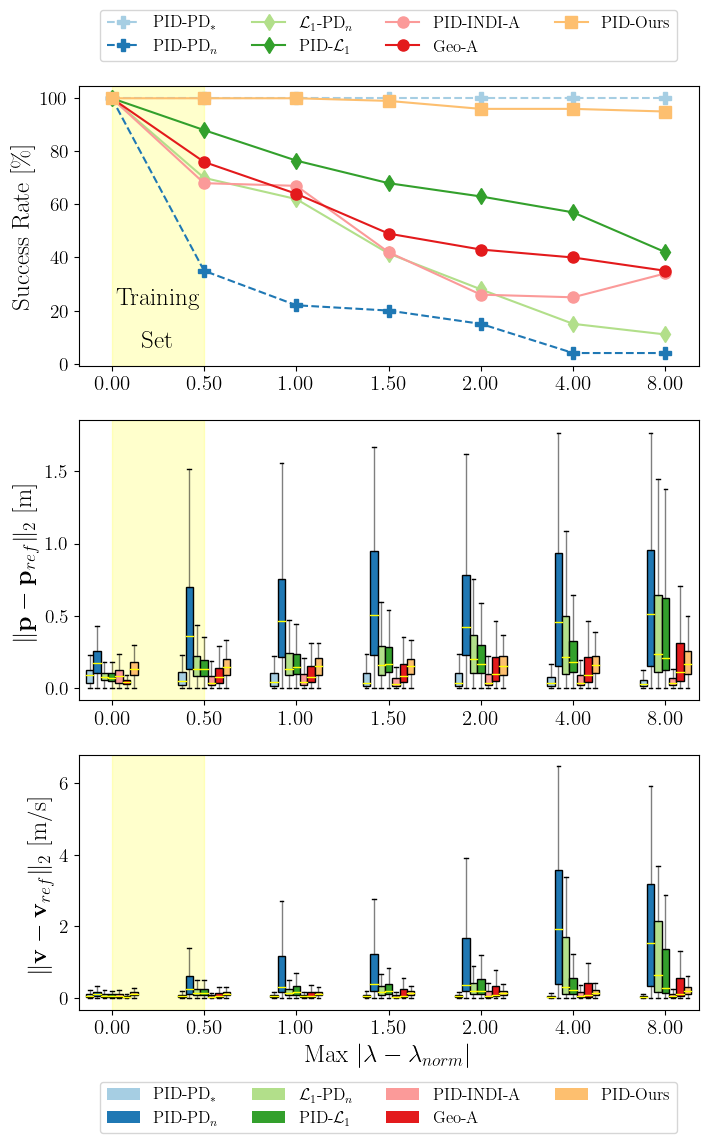}
\caption{\rebuttal{We evaluate the performance of our method and all baselines on extended quadrotor parameters range unseen at training time. We use metrics $\delta= \max |\lambda-\lambda_{\mathrm{norm}}|$, the maximum difference of all sampled quadcopters away from the nominal model, to define the quadrotor randomization range. We plot \textbf{Top}: the success rate, \textbf{Middle}: the box plot of the position tracking error and \textbf{Bottom}: the box plot of the position tracking error of our method and all baselines over the parameter randomization range. At each data point, the result is calculated over 100 experiments. All sampled quadcopters within the shaded area between 0 and 0.5 belong to the training range of Table~\ref{tab:randomization}. Note that for better visualization, the x-axis is not to scale.}}
\vspace{-3ex}
\label{fig:generlaization-plots}
\end{figure}



We evaluate the task of tracking trajectories on held-out quadrotor parameter range. 
In particular, we aim to determine the extent to which deviations from the nominal model cause our controller and other baseline controllers to fail.
We use the same baselines as in previous sections, with the nominal model $\lambda_{\mathrm{norm}}$ now obtained at the mid-point of the training range. 
For ease of representation, we express $\lambda$ and $\lambda_{\mathrm{norm}}$ in a numeric way, with its value equal to the scaling constant $c$. Therefore, $\lambda_{\mathrm{norm}}=0.5$ and $\lambda \in [0,1]$ is the training set of Table~\ref{tab:randomization}.

We use the metrics 
\begin{align}    
\delta = \max |\lambda-\lambda_{\mathrm{norm}}|
\end{align}
to define the extent of the range of sampled quadrotor parameters. 
In particular, when $\delta = 0$, the sampled quadrotor $\lambda$ is the nominal model $\lambda_{\mathrm{norm}}$ with noises; when $\delta=0.5$, $\lambda \in [0, 1]$ is the training set in Table~\ref{tab:randomization}. 
We extend $\delta$ up to 8 to evaluate the task of trajectory tracking for randomly sampled quadrotors, in which the sampling range is 16 times wider than the training set. 
Figure~\ref{fig:train-test-visual} provides a visualization of the differences between the two sets. We randomly sample 2,000 quadcopters within the $\delta=8$ range and within the training range, plotting them as scatter points based on arm length, mass, and torque-to-thrust coefficient. This illustration highlights the differences in size, mass, and motor strength between the training and testing sets. It is evident that the testing set is significantly outside the training distribution

%
%
The success rate and the position and velocity tracking error distribution are reported in Figure~\ref{fig:generlaization-plots}.
%
%
Our method achieves near 100\% success rate until $\delta =8$ where it drops to 95\%. In contrast, all other baselines, except for \textit{PID-PD$_*$}, exhibit significant performance degradation as the model mismatch from the nominal model increases.
In addition, the average position tracking error at $\delta=8$ of our method is still close to that at $\delta=0$, with only 37.0\% increase. Compared to the strongest baseline \textit{PID-\lone} in terms of success rate in Section~\ref{sec:sim-baseline-results} whose position tracking error has grown by 483.7\% compared to that at the nominal model.
%


\section{Hardware Experiments}

\begin{figure}[t]
\centering
\includegraphics[width=\linewidth]{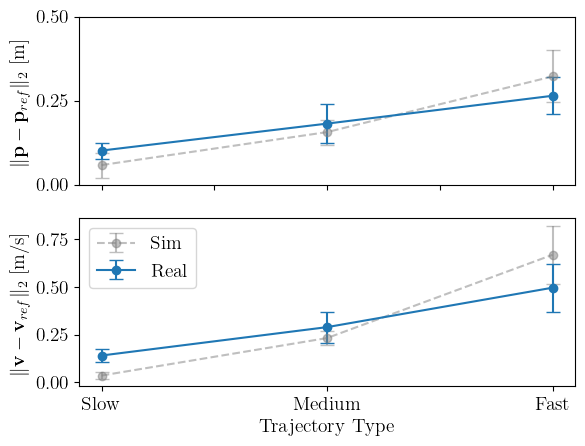}
\caption{To evaluate the sim-to-real gap, we control the large quadrotor with our proposed controller along circular trajectories at 3 speed settings: \textit{Slow}, \textit{Medium} and \textit{Fast}. The figure illustrates the error bars for the distribution of position and velocity tracking errors. Both plots exhibit similar trends and magnitudes, suggesting that our simulator effectively reflects real-world dynamics and thus supports the reliability of our results. These results are derived from 10 simulated flights and 3 real-world flights.
}
\vspace{-3ex}
\label{fig:sim2real-correlate}
\end{figure}

In this section, we transition from simulation to hardware experiments. We first investigate the sim-to-real correlation to ensure the validity of our simulation results. Subsequently, we conduct a comparative analysis on disturbance rejection tasks, comparing our method against the best baseline identified in our simulation tests.

\subsection{Sim-to-Real Correlation}

We validate our simulation results through two sets of hardware and simulation experiments to examine the simulation-to-reality gap. 
We fly the large quadrotor along circular trajectories with our control framework \textit{PID-Ours} at 3 speed settings: \textit{Slow}, \textit{Medium} and \textit{Fast}. Subsequently, we simulate the same flight paths of the same vehicle with our method again using the Flightmare simulator~\cite{song2020flightmare}.  
The trajectories involve circling with a 1-meter radius with completion in different durations: 8s (\textit{Slow}), 4s (\textit{Medium}), and 3s (\textit{Fast}).
The results, illustrated in Figure~\ref{fig:sim2real-correlate}, show the distribution of position and velocity tracking errors for both sets of experiments across all speed settings. 
We also compute the Pearson Correlation Coefficient~\cite{freedman2007statistics,benesty2009pearson}, which shows a moderate positive correlation (0.652) between the position tracking errors in the simulation and real-world tests, with a statistically significant P-value of 0.002. 
These findings suggest that our simulator captures the dynamics of the real world reasonably well, supporting the reliability of our simulation results.

\subsection{Baseline Comparison}

\begin{table*}[t]
\caption{\label{tab:lowlevel-hw} We compare the performance of our controller PID-Ours to the best baseline PID-PD$_*$ controller that uses accurate model information in Table~\ref{tab:highlevel}, in tasks of disturbance rejections that are hard to replicate in the simulation. Since the two control frameworks share the same high-level controller, \rebuttal{we measure performance on tracking high-level commands and other metrics used in simulation, i.e., maximum position error, position, and velocity errors.} The comparison is run on the large quadcopter and the small quadcopter. We compare these approaches' performance tracking a circular trajectory at \textit{Medium} speed under 3 tasks. \textbf{Disturb Free}: track the trajectory without any disturbances. \textbf{Off-center Payload}: track the trajectory under an unknown off-center payload. \textbf{Wind}: track the trajectory under wind. We also evaluate their performance on an additional task, \textbf{Thrust Loss}: take off and hover with one single motor experiencing 20\% thrust loss. \rebuttal{Note that the maximum position error is not reported for this task because it involves a setpoint rather than a trajectory; hence, the maximum position error would be the position error at takeoff.} The results are from 3 experiments for each method per task.}
\setlength{\tabcolsep}{3pt}
\def\arraystretch{1.2}
\centering
\begin{tabular*}{0.75\textwidth}{@{\extracolsep{\fill}}lccccccc}
\toprule
 & \multirow{2}{*}{\begin{tabular}[c]{@{}c@{}}Vehicle\end{tabular}}  & Low-level & Thrust  & Average Angular  & \rebuttal{Max Pos.} & \rebuttal{Position} & \rebuttal{Velocity} \\
 &&Controller&RMSE (m/s$^2$)&RMSE (rad/s)&\rebuttal{Err (m)}&\rebuttal{RMSE (m)}&\rebuttal{RMSE (m/s)}\\
\midrule
\multirow{4}{*}{\begin{tabular}[c]{@{}c@{}}\textbf{Disturb Free}\end{tabular}} 
 & \multirow{2}{*}{small} & PD$_*$ & \textbf{0.132} & \textbf{0.339} &\rebuttal{\textbf{0.094}}& \rebuttal{\textbf{0.017}} & \rebuttal{\textbf{0.022}} \\ 
 &  & Ours & 0.280 & 0.721 &\rebuttal{0.098}& \rebuttal{0.033} & \rebuttal{0.047} \\ \cline{2-8} 
 & \multirow{2}{*}{large} & PD$_*$ & \textbf{2.327} & \textbf{0.296} &\rebuttal{\textbf{0.108}}& \rebuttal{\textbf{0.046}} & \rebuttal{\textbf{0.132}} \\ 
 &  & Ours & 3.325 & 0.413 & \rebuttal{0.122}&\rebuttal{0.057} & \rebuttal{0.197} \\
\midrule
\multirow{4}{*}{\textbf{Wind}} 
 & \multirow{2}{*}{small} & PD$_*$ & 2.477 & 1.365 &\rebuttal{0.245}&\rebuttal{0.175}  & \rebuttal{0.328} \\
 &  & Ours & \textbf{1.659} & \textbf{0.429} & \rebuttal{\textbf{0.191}} & \rebuttal{\textbf{0.130}} & \rebuttal{\textbf{0.205}} \\ \cline{2-8} 
 & \multirow{2}{*}{large} & PD$_*$ & \textbf{2.916} & 0.543 &\rebuttal{0.576}& \rebuttal{\textbf{0.072}} & \rebuttal{0.188} \\
 &  & Ours & 3.549 & \textbf{0.523} & \rebuttal{\textbf{0.341}}&\rebuttal{0.076} & \rebuttal{\textbf{0.131}} \\
\midrule
\multirow{4}{*}{\begin{tabular}[c]{@{}c@{}}\textbf{Off-center Payload}\end{tabular}} 
 & \multirow{2}{*}{small} & PD$_*$ & 0.730 & 1.360 & \rebuttal{0.435}&\rebuttal{0.260} & \rebuttal{0.323} \\ 
 &  & Ours & \textbf{0.679} & \textbf{0.510} & \rebuttal{\textbf{0.185}}&\rebuttal{\textbf{0.130}} & \rebuttal{\textbf{0.158}} \\ \cline{2-8} 
 & \multirow{2}{*}{large} & PD$_*$ & 3.420 & 0.697 & \rebuttal{0.703}&\rebuttal{0.404} & \rebuttal{0.564}\\ 
 &  & Ours & \textbf{2.808} & \textbf{0.456} & \rebuttal{\textbf{0.253}}&\rebuttal{\textbf{0.167}} & \rebuttal{\textbf{0.215}} \\
\midrule
\multirow{4}{*}{\begin{tabular}[c]{@{}c@{}}\textbf{Thrust Loss}\end{tabular}} 
 & \multirow{2}{*}{small} & PD$_*$ & 0.689 & 1.408 & \rebuttal{N.A.}&\rebuttal{0.470} & \rebuttal{0.389} \\
 &  & Ours & \textbf{0.403} & \textbf{0.576} & \rebuttal{N.A.}& \rebuttal{\textbf{0.325}} & \rebuttal{\textbf{0.196}} \\ \cline{2-8}
 & \multirow{2}{*}{large} & PD$_*$ & Fail & Fail & \rebuttal{Fail} & \rebuttal{Fail} & \rebuttal{Fail} \\
 &  & Ours & \textbf{2.167} & \textbf{0.355} & \rebuttal{N.A.}&\rebuttal{\textbf{0.572}} & \rebuttal{\textbf{0.224}} \\
\bottomrule
\end{tabular*}
\end{table*}

\begin{figure}[t]
    \centering
    \includegraphics[width=\linewidth]{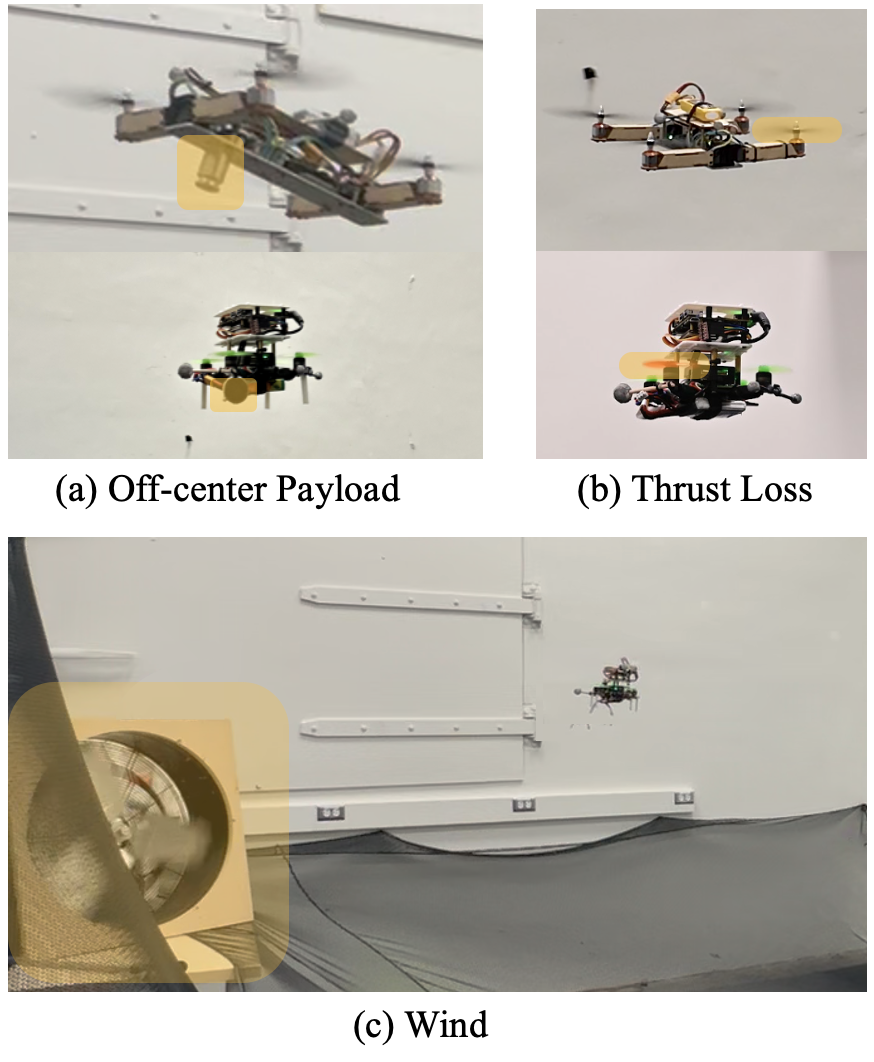}
    \caption{\textbf{(a)} Large quadrotor and small quadrotor mounted with an off-center payload. For the large vehicle, we mount a 200g payload to the farthest end of the body frame from the center of gravity. For the small one, we mount a 30g payload directly under one of its motors. \textbf{(b)} Both quadcopters experiencing a 20\% thrust loss from one of its actuator, which is achieved by hard-coding the firmware code. \textbf{(c)} Small quadrotor tracking a circular trajectory under wind up to 3.5m/s. Large quadrotor undergoes this experiment with the same setup.}
    \label{fig:hardware-disturbance}
\end{figure}

We test our approach in the physical world and compare its performance to the \textit{PID-PD$_*$} controller that has access to the platform's parameters and has been specifically tuned to the platform with in-flight tests.
In contrast, our approach has no knowledge whatsoever of the physical characteristics of the system and does not calibrate or fine-tune with real-world flight data.
Our hardware experiments are designed to evaluate our method's capability to handle disturbances that are challenging to simulate.
We test on the task of tracking a 1m circular trajectory with completion duration of 6s without any disturbances, under an off-center payload up to 20\% of body mass which is attached to the farthest end on the body frame of the tested quadcopter, and under wind up to 3.5m/s.
We also test on the task of taking off and hovering with one single motor experiencing 20\% thrust loss. This is achieved by modifying the quadcopter's firmware to hard-code the hardware command sent to the affected motor so that the thrust produced by this particular actuator is always 80\% of its desired value. We change the firmware to mimic the partial failure in the system in a controlled manner, instead of intentionally crashing the vehicle, to avoid actual damage and facilitate reproducibility. All experiment setup details are shown in Figure~\ref{fig:hardware-disturbance}.

The high-level controller for both methods is a PID controller, the same as in Section~\ref{sec:ex-setup}. 
The only variation in the two control frameworks is the low-level controller. 
%
\rebuttal{We asses performance using the same metrics of the simulation experiments. However, to test the controller's ability to track desired commands, we additionally measure the average tracking error of (i) mass-normalized thrust and (ii) angular velocity based on the high-level controller's commands.}
We define a failure as a situation where the human operator has to intervene to prevent the quadcopter from crashing. 
The results of these experiments are reported in table~\ref{tab:lowlevel-hw}.
Note that in nearly all experiments, the thrust tracking RMSE for large quadrotor is much higher than that for the small quadrotor. This discrepancy arises because the large quadrotor has more powerful actuators, which can induce greater vibrations in the system, subsequently affecting the accelerometer readings. Therefore, this difference does not necessarily indicate that the thrust tracking for the large quadrotor is worse than for the small quadrotor. It is more appropriate to compare the performance of different methods within each platform, rather than across them.

Our approach and the \textit{PD$_*$} baseline perform similarly in disturbance-free experiments, with the \textit{PD$_*$} controller slightly outperforming ours.  
The latter difference in performance is justified since the \textit{PD$_*$} controller is specifically tuned for each quadcopter.
Our method significantly outperforms the model-based \textit{PD$_*$} in both metrics in the presence of off-center payload and thrust loss. In particular, \textit{PID-PD$_*$} experiences a total failure in the case of thrust loss on the large quadrotor platform. 
Both disturbances create a large model mismatch from the nominal model that the \textit{PID-PD$_*$} uses. 
Our method is able to adapt to the mismatch well with a similar high-level command tracking error compared to that at the disturbance-free case. Conversely, the \textit{PID-PD$_*$} controller is not as adaptive. Indeed, its tracking error is up to 4.53 times higher for thrust tracking and up to 3.15 times for angular velocity tracking. 
Finally, the purpose of wind experiments is
to evaluate our controller’s performance under non-constant disturbances. Note that such time-varying disturbances were not present during training. 
Our controller is robust to wind disturbances with a comparable tracking performance as the \textit{PID-PD$_*$} on the large quadrotor and a significantly smaller tracking error on the small quadrotor. 
The small size and weight of this platform make it more susceptible to the interaction of the wind with its body and rotors, which can alter its aerodynamic properties, such as affecting the effective angle of attack on the rotors. 
Therefore, the model mismatch in terms of alteration in aerodynamic properties is more prominent on the small quadrotor than on the large quadrotor. Our controller can adapt well to such disturbances.
\rebuttal{Across all sets of experiments, position and velocity errors correlate with high-level tracking command errors, as larger high-level errors lead to compounded position and velocity inaccuracies. This further emphasizes the importance of robustness and adaptivity of the low-level controller.}

\rebuttal{\subsection{Adaptation Analysis}}
\begin{figure}[t]
    \centering
    \includegraphics[width=\linewidth]{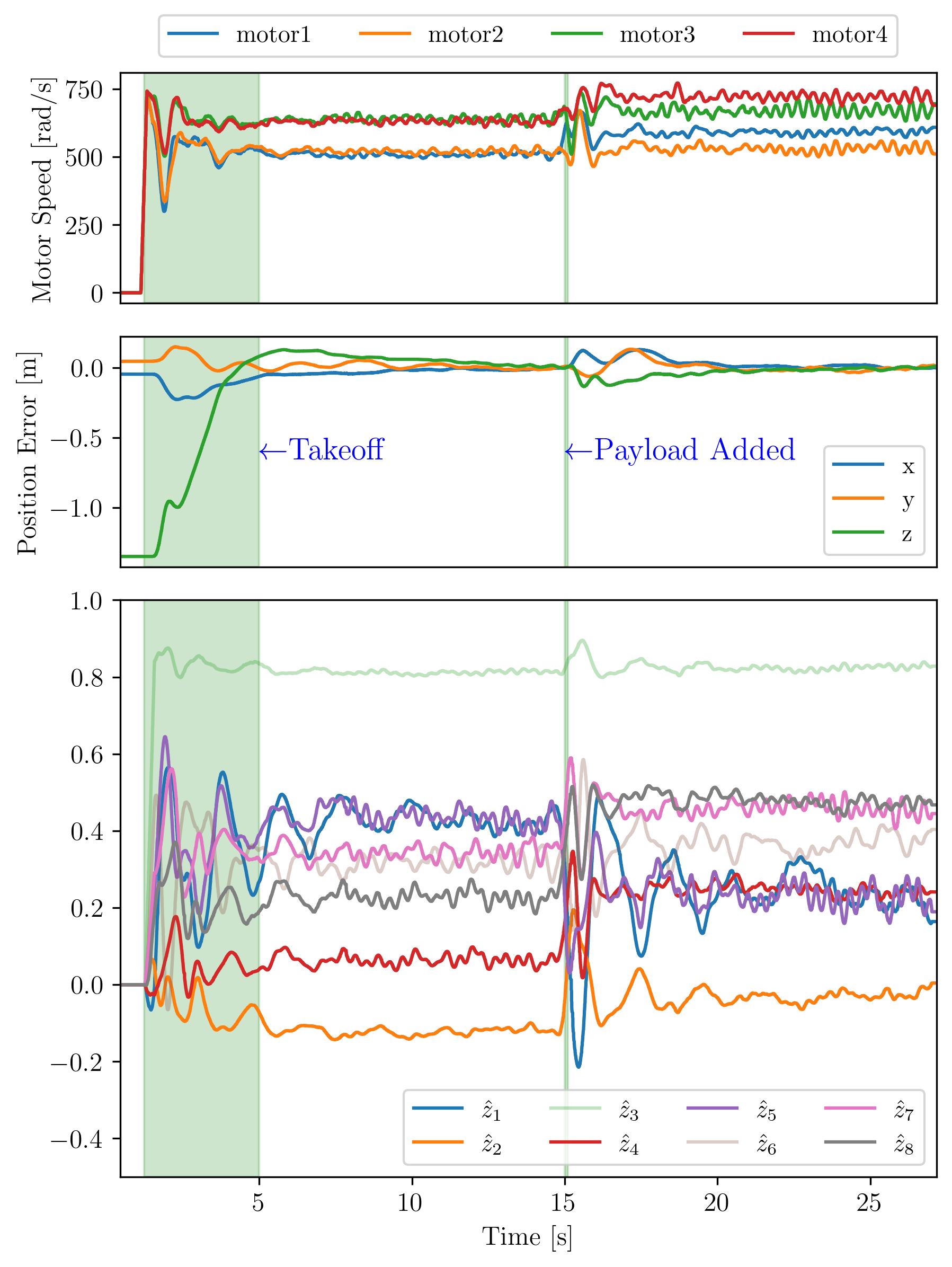}
    \caption{\rebuttal{Visualization of the off-center payload experiment on the large quadrotor. The plots show: \textbf{Top}: motor speeds commanded by our low-level controller, \textbf{Middle}: absolute position tracking throughout the experiment, and \textbf{Bottom}: each element of the estimated 8-dim intrinsics $\hat{\bm{z}}_t$. The timeline spans from takeoff to approximately 10 seconds after the payload is added. Shaded regions indicate phase transitions, with annotations indicating key events. In the bottom plot, elements of $\hat{\bm{z}}_t$ that exhibit more than a 10\% change in their average value before and after payload attachment are highlighted, showing the components most affected by the payload disturbance.}}
    \label{fig:real-hover-payload-vis}
\end{figure}

\rebuttal{We consider the controller performance for an off-center payload, using the large quadcopter. The result is shown in Figure~\ref{fig:real-hover-payload-vis}.}
%
\rebuttal{During takeoff, all components of $\bm{\hat{z}}$ exhibit variations as the quadrotor stabilizes. The vehicle transitions from stationary on the ground to takeoff at the commanded velocity within approximately 0.2 seconds, aligning with the adaptation module’s time window (100 state-action pairs at 500Hz). This adaptation process is significantly faster than approaches that rely on online system identification~\cite{wuest2019online,svacha2020imuest}, which typically require 10–15 seconds of flight to converge to an accurate model estimate.}

\rebuttal{When the payload is added, multiple components undergo a sharp transient response, indicating an update in the controller’s internal representation. The highlighted components show a sustained shift, suggesting that these dimensions capture key physical properties such as mass distribution and inertial variations. In contrast, the average value of $\bm{\hat{z}}_3$ and $\bm{\hat{z}}_6$ remains largely unchanged after convergence, suggesting that these dimensions encode properties that are unaffected by an off-center payload, such as lateral forces or yaw torque.}

\rebuttal{\section{Ablation Study}}
\rebuttal{Training a low-level controller that adapts across different quadrotors involves various complexities. This section presents an ablation study analyzing the training curves to evaluate the impact of the IL component and reward design. Additionally, we examine how the RL and IL components, and our control-level selection, influence performance.}

\rebuttal{\subsection{Training Curve Analysis}\label{sec:il-rationale}}

\subsubsection{Integration of IL}

\begin{figure}[t]
    \centering
    \includegraphics[width=\linewidth]{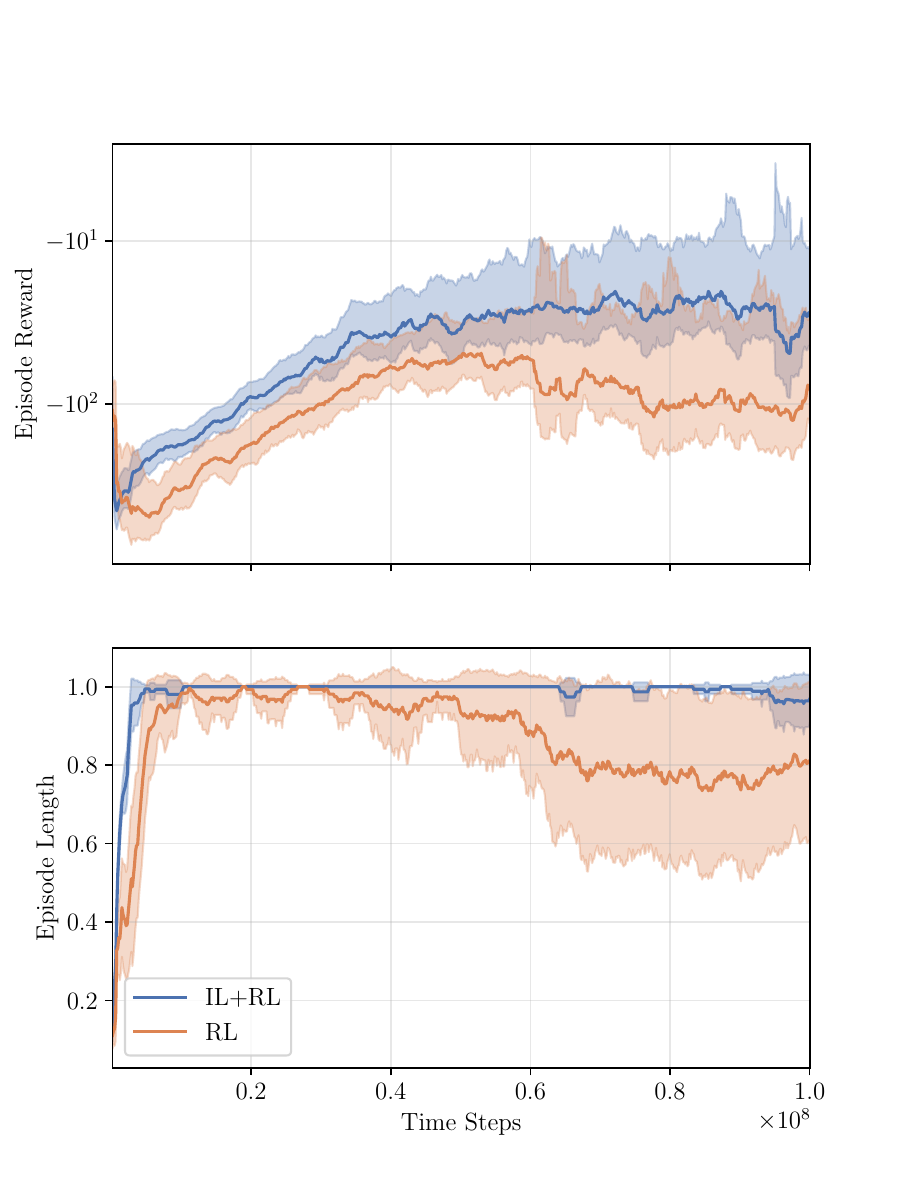}
    \caption{\rebuttal{Comparison of episode rewards (top) and episode lengths (bottom) for policies trained using both IL and RL, versus RL only, across 10 consecutive random seeds (with the bold line representing the mean and the shaded area representing the 95\% confidence interval). The episode length is normalized by dividing the vehicle's surviving duration by the maximum episode duration. An episode length of 1 indicates that the vehicle controlled by the trained policy survives for the entire episode without crashing. For better visualization, we plot the reward curve in log scale.}}
    \label{fig:IL_RL_reward}
\end{figure}

\begin{table}
\caption{\label{tab:IL_RL_comp}\rebuttal{Comparison of average episode reward and length with 95\% confidence interval (CI) over the last 10\% of the training process for IL+RL and RL-only policies.}}
\centering
\rebuttal{ 
\begin{tabular*}{0.4\textwidth}{@{}lcc@{}}
\toprule
    & Mean Episode Reward & Mean Episode Length \\
    & $\pm$95\% CI & $\pm$95\% CI \\
\midrule
IL+RL  & \textbf{-31.72$\pm$22.02} & \textbf{0.97$\pm$0.05}\\
RL & -102.55$\pm$76.39 & 0.79$\pm$0.20\\
\bottomrule
\end{tabular*}
}
\end{table}

\rebuttal{We adopt a dual strategy which combines IL and RL and adaptively adjust the relative weight between the two losses. This learning approach improves the training of the low-level controller compared to training solely with RL, as in our previous conference paper~\cite{zhang2023universal}. Figure~\ref{fig:IL_RL_reward} presents an ablation of the IL component of the loss. We also report the average episode reward and length with 95\% confidence interval (CI) over the last 10\% of training process in Table~\ref{tab:IL_RL_comp}. 
The policy trained with both IL and RL exhibits steady improvement in both reward and episode length throughout the learning phase, maintaining an episode length close to 1 after approximately 10M steps. 
In contrast, the RL-only baseline shows less consistent performance with a larger variance. Although its episode length approaches 1 around 20M steps, it soon begins to diverge, indicating instability. Additionally, the RL-only baseline maintains a consistently lower average reward compared to the IL-and-RL method, highlighting inferior tracking performance. As training progresses, the instability worsens, leading to a decline in reward. To further analyze these components, we conduct an ablation study in the context of our simulation experiments and evaluate the roles of the RL and IL components in Section~\ref{sec:ablation-sim-exp}.}

\subsubsection{Reward Design}

\begin{figure}
    \centering
    \includegraphics[width=\linewidth]{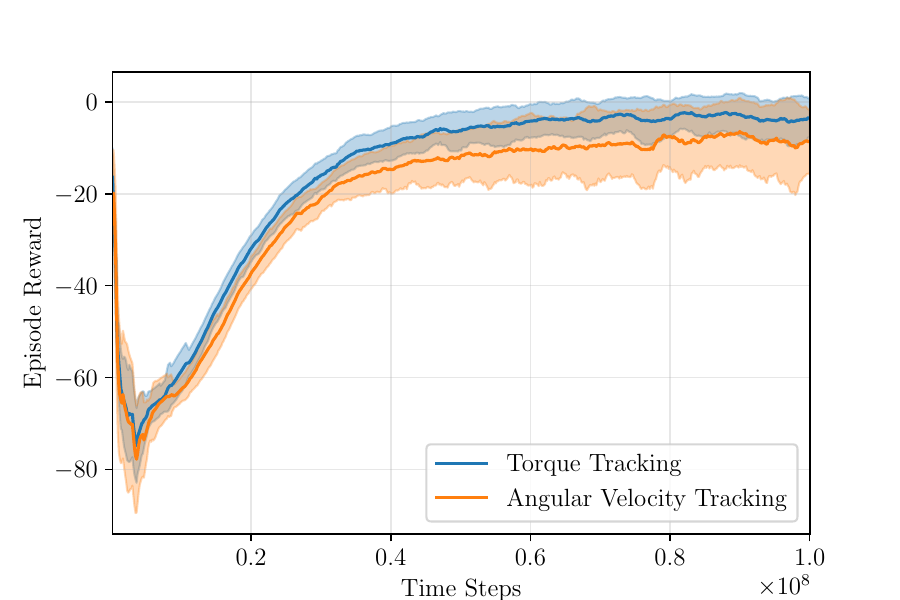}
    \caption{\rebuttal{Comparison of episode reward for policies trained with torque tracking, versus with angular velocity tracking, across 10 consecutive random seeds (with the bold line representing the mean and the shaded area representing the 95\% confidence interval. The episode reward does not include the torque/angular velocity reward term as they are the variation in reward design.}}
    \label{fig:torque_angVel_plot}
\end{figure}

\begin{table}[]
\caption{\label{tab:angRew_ablation_comp}\rebuttal{Comparison of average episode reward with 95\% confidence interval (CI) over the last 10\% of the training process for policies with torque tracking and with angular velocity tracking. The episode reward does not include the torque or angular velocity tracking reward for comparability.}}
\centering
\rebuttal{ 
\begin{tabular*}{0.35\textwidth}{@{}lcc@{}}
\toprule
    & Mean Episode Reward \\
    & $\pm$ 95\% CI  \\
\midrule
Torque Tracking  & \textbf{-3.91$\pm$4.67}\\
Angular Velocity Tracking & -8.74$\pm$ 8.15 \\
\bottomrule
\end{tabular*}
}
\end{table}

\rebuttal{We reward torque tracking instead of angular velocity tracking, as described in Section~\ref{sec:reward-des}. While angular velocity is a high-level command and a key observation, our experiments show that torque tracking leads to better sim-to-real transfer.}

\rebuttal{To support this, we compare policies trained with torque tracking and angular velocity tracking in Figure~\ref{fig:torque_angVel_plot}, keeping all other reward components and hyperparameters identical. Table~\ref{tab:angRew_ablation_comp} reports the average reward over the last 10\% of training. The torque-tracking policy exhibits stable improvement throughout training, whereas the angular velocity-tracking policy achieves lower rewards and shows greater fluctuations.
%
Such behavior not only increases oscillation penalties but also degrades sim-to-real transfer, as saturated commands can overheat motors and cause hardware damage.}
\rebuttal{Torque tracking mitigates these issues by providing a more immediate and direct learning signal. Since torque responds directly to commanded motor speed, it enables faster corrections, which is particularly important for low-level controllers operating at 500Hz. By contrast, angular velocity is derived through integration, making its feedback less immediate for RL.}

\rebuttal{\subsection{Simulation Performance}\label{sec:ablation-sim-exp}}

\begin{figure}[t]
\centering
\includegraphics[width=\linewidth]{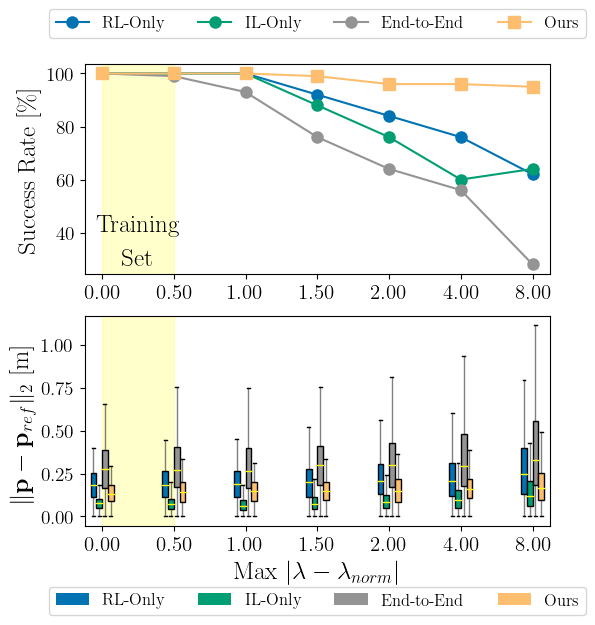}
\caption{\rebuttal{We perform an ablation study under the same experimental conditions as Figure~\ref{fig:generlaization-plots}, comparing our method (\textit{Ours}) against three variations: (i) \textit{RL-Only}, which removes the IL component and trains solely via reinforcement learning, (ii) \textit{IL-Only}, which trains the controller exclusively through imitation learning using DAgger~\cite{ross2011dagger}, and (iii) \textit{End-to-End}, a single learned policy with our method but integrating high-level and low-level controllers. The results of \textit{Ours} are taken directly from Figure~\ref{fig:generlaization-plots}. The results are shown for \textbf{Top}: success rate, and \textbf{Bottom}: the position tracking error (box plots). At each data point, the result is calculated over 100 experiments. All sampled quadcopters within the gray shaded area belong to the training range of Table~\ref{tab:randomization}. Note that for better visualization, the x-axis is not to scale.}}
\vspace{-4ex}
\label{fig:ablation-sim-plots}
\end{figure}

\rebuttal{In this section, we present an ablation study to investigate the contribution of individual components in our method design within the context of our simulation experiments. Specifically, we compare our method against three variations: (i) \textit{RL-Only}, which excludes the IL component and relies entirely on reinforcement learning for training, (ii) \textit{IL-Only}, which trains the controller solely through imitation learning using DAgger~\cite{ross2011dagger} without separation of networks and phases as in our framework, and (iii) \textit{End-to-End}, which replaces our hierarchical control structure with a unified policy that directly fuses high-level and low-level control, trained using our method. The first two variants provide insight into the significance of individual training components, extending the analysis from the training curves in Section~\ref{sec:il-rationale}. The third variation focuses on justifying our design choice for separate control levels.}
\rebuttal{We perform the experiment on the same generalization task described in Section~\ref{sec:generalization}. The success rate and the distribution of position tracking errors are presented in Figure~\ref{fig:ablation-sim-plots}. Note that the results for \textit{Ours} are the same as shown in Figure~\ref{fig:generlaization-plots}.}

\subsubsection{RL and IL Ablation}
\rebuttal{Both \textit{RL-Only} and \textit{IL-Only} experience a decline in success rate as the model mismatch increases beyond the training set, ultimately dropping to around 60\%. \textit{RL-Only} demonstrates a slightly higher success rate than \textit{IL-Only}. In terms of tracking error, \textit{IL-Only} achieves the smallest distribution among the methods. However, both approaches exhibit increased tracking error as the model mismatch grows. At $\delta=8$, the average tracking error for \textit{IL-Only} increases by 92.4\%, while for \textit{RL-Only}, it grows by 61.4\%, compared to their respective values at $\delta=0$.
In contrast, our method, which combines RL and IL, maintains a near 100\% success rate across all levels of model mismatch while achieving consistent tracking performance. This raises the question: how can the combination of two methods, each with limited generalization capability, yield better results?}

\rebuttal{We hypothesize that it is because IL and RL optimize distinct objectives. IL focuses on minimizing instantaneous tracking error, while RL targets both tracking error and success rate. RL achieves this dual optimization by incorporating a reward signal for tracking error, as described in Section~\ref{sec:reward-des}, and utilizing episode termination conditions to influence success rate. As a result, RL may accept higher tracking errors to extend episode length, thereby improving success rates. This explains why \textit{RL-Only} outperforms \textit{IL-Only} in terms of success rate.
When combined, RL can prioritize success rate optimization, as the IL loss naturally handles tracking error minimization. This complementary relationship allows the two methods to reinforce each other, resulting in improved overall performance. Future work could explore this hypothesis further, investigating the underlying mechanisms and conditions that enable a synergy between IL and RL.}

\subsubsection{End-to-End Evaluation}

\rebuttal{\textit{End-to-End} is trained using our proposed method but as a unified policy that combines high-level and low-level control. Its input includes the same features as our policy, along with additional information: position ($\mathbb{R}^3$), velocity ($\mathbb{R}^3$), and the difference between the current and goal positions ($\mathbb{R}^3$). Its output, like our policy, is motor speed, which motivates the name \textit{End-to-End}.
Despite being trained with the same method and without ablations, \textit{End-to-End} demonstrates the lowest success rate and the largest average tracking error among all tested variations. In contrast, the other variations (\textit{RL-Only}, \textit{IL-Only} and \textit{Ours}) all utilize a low-level controller. The failure of End-to-End learning likely comes from the challenge of jointly encoding both high-level trajectory tracking and low-level motor adaptation within a single policy. Maintaining a modular structure allows each level to focus on distinct aspects of control, facilitating generalization beyond the training set.}

\rebuttal{Moreover, when compared to state-of-the-art model-based methods in Figure~\ref{fig:generlaization-plots}, its performance is similarly poor. At $\delta=8$, \textit{End-to-End} achieves a success rate of only 28\%, which is marginally better than \textit{PID-PD$_n$} and \textit{\lone-PD$_n$}, both of which lack adaptation in the low-level controller. This comparison in parallel also highlights the importance of separating control levels and employing an adaptive low-level controller to handle the large model disparities defined in our problem.}
\section{Conclusion}
This work demonstrates how a single adaptive controller can effectively bridge the gap between high-level planning and the intricate physical dynamics by adapting to model disparities between quadcopters down to the motor level. Our design focuses on creating a low-level controller intended to replace traditional low-level quadcopter controllers, thereby eliminating the need for accurate model estimation and iterative parameter tuning. Our approach leverages a combination of imitation learning from model-based controllers and reinforcement learning to address the challenges of training a sensor-to-actuator controller at high frequencies. The introduction of an instant reward feedback ensures that the controller remains responsive and agile. In addition, we develop a quadcopter randomization method during training that aligns with real-world constraints, further enhancing its adaptability. 
The controller's ability to estimate a latent representation of system parameters from sensor-action history, along with realistic domain randomization, empowers it to generalize across a broad spectrum of quadcopter dynamics. This capability extends to unseen parameters, with an adaptation range up to 16 times broader than the training set. The single policy trained solely in simulation can be deployed zero-shot to real-world quadcopters with vastly different designs and hardware characteristics. It also demonstrates rapid adaptation to unknown disturbances, such as off-center payloads, wind, and loss of efficiency in motors.
These results highlight the potential of our approach for extreme adaptation for drones and other robotic systems, while enabling robust control in the face of real-world uncertainties.

\section{Acknowledgement}
This work was supported by the Hong Kong Center for Logistics Robotics, the DARPA Transfer from Imprecise and Abstract Models to Autonomous Technologies (TIAMAT) program, the Graduate Division Block Grant of the Dept. of Mechanical Engineering, UC Berkeley, and ONR MURI award N00014-21-1-2801.
The authors would like to thank Ruiqi Zhang and Teaya Yang for their help with the experiments. \rebuttal{The authors also would like to thank Daniel Gehrig for his valuable input in the ablation analysis of our method. }
The experimental testbed at the HiPeRLab is the result of contributions of many people, a full list of which can be found at \url{hiperlab.berkeley.edu/members/}.
\appendices
\section{Baseline Implementation}\label{app:baseline-impl}
A detailed explanation of each baseline's implementation is provided to support reproducibility. The baselines are structured into high-level and low-level controllers. Most baseline names follow the format \textit{high-level controller}-\textit{low-level controller}, with exceptions explicitly noted. For instance, \textit{PID-PD$_*$} indicates that PID serves as the high-level controller, while PD$_*$ represents the low-level controller. Identical control names imply the same implementation, and explanations are omitted if they were already described for previous baselines.

\subsection{PID-PD$_*$}
\subsubsection{PID}
The high-level PID controller is implemented as described in Section~\ref{sec:ex-setup}. It generates high-level commands, including body rates and mass-normalized collective thrust, for other low-level controllers to track.

\subsubsection{PD$_*$}
This baseline matches the expert controller used during training, as outlined in Section~\ref{sec:base_policy}. Unlike other baselines, it has access to ground-truth model parameters for control calculations. Consequently, it serves as both an expert reference and a performance upper bound, providing context for evaluation.

\subsection{PID-PD$_n$}
\subsubsection{PD$_n$}
This baseline is similar to PD$_*$ but calculates control using only the nominal model $\lambda_{\mathrm{norm}}$. As such, it represents a performance lower bound for comparison.

\subsection{PID-\lone} \label{app:l1-low} 
\begin{figure}
    \centering
    \includegraphics[width=\linewidth]{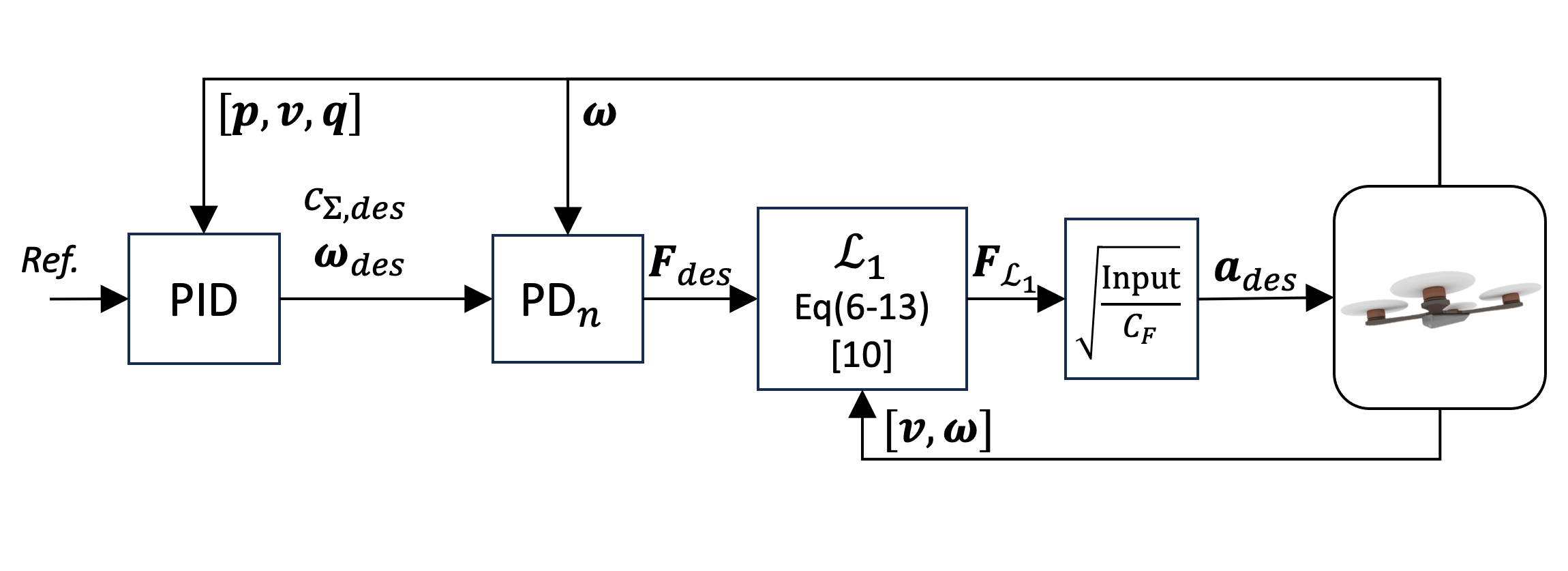}
    \caption{\rebuttal{The control diagram of \textit{PID-\lone}}.}
    \label{fig:l1-low-level}
\end{figure}
\subsubsection{\lonesp at Low Level}

We implement the \lonesp adaptive controller as an augmentation to the \textit{PID-PD$_n$} control, applying the adaptive controller at the low level. Beyond serving as a baseline, it also acts as a counterpart to \textit{\lone-PD$_n$}, which applies the same adaptation within the high-level control structure. The comparison between the two methods can validate our design choice of control hierarchy. 
The implementation is same to~\cite{hanover2021performance} and
%
%
the controller's block diagram is shown in Figure~\ref{fig:l1-low-level}. The \lonesp control receives the desired motor thrust from PD$_n$ and computes the augmented thrust using the control and adaptation law defined in (6-13) in~\cite{hanover2021performance}. The resulting thrust command is then converted into motor speed. Unless otherwise specified, all model constants used for control calculations in this framework and in all baselines, except PD$_*$, are derived from the nominal model.

\subsection{\lone-PD$_n$}
\begin{figure}
    \centering
    \includegraphics[width=\linewidth]{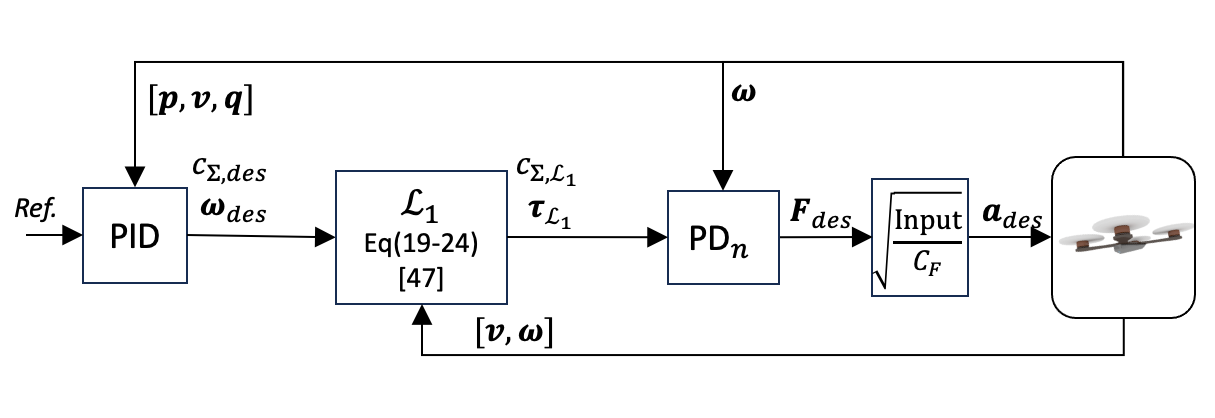}
    \caption{\rebuttal{The control diagram of \textit{\lone-PD$_n$}}}
    \label{fig:l1-high-level}
\end{figure}
\subsubsection{\lonesp at High Level} We implement a similar \lonesp controller as described previously but the uncertainty is compensated at the level of thrust and body torque. We refer to the framework in~\cite{pravitra2020l1torque} in which the \lonesp controller gives thrust and torque command. The control and adaptation law is defined in (19-24) in~\cite{pravitra2020l1torque}. Figure~\ref{fig:l1-high-level} outlines the block diagram of the controller. 

\subsection{Geo-A}
\begin{figure}
    \centering
    \includegraphics[width=\linewidth]{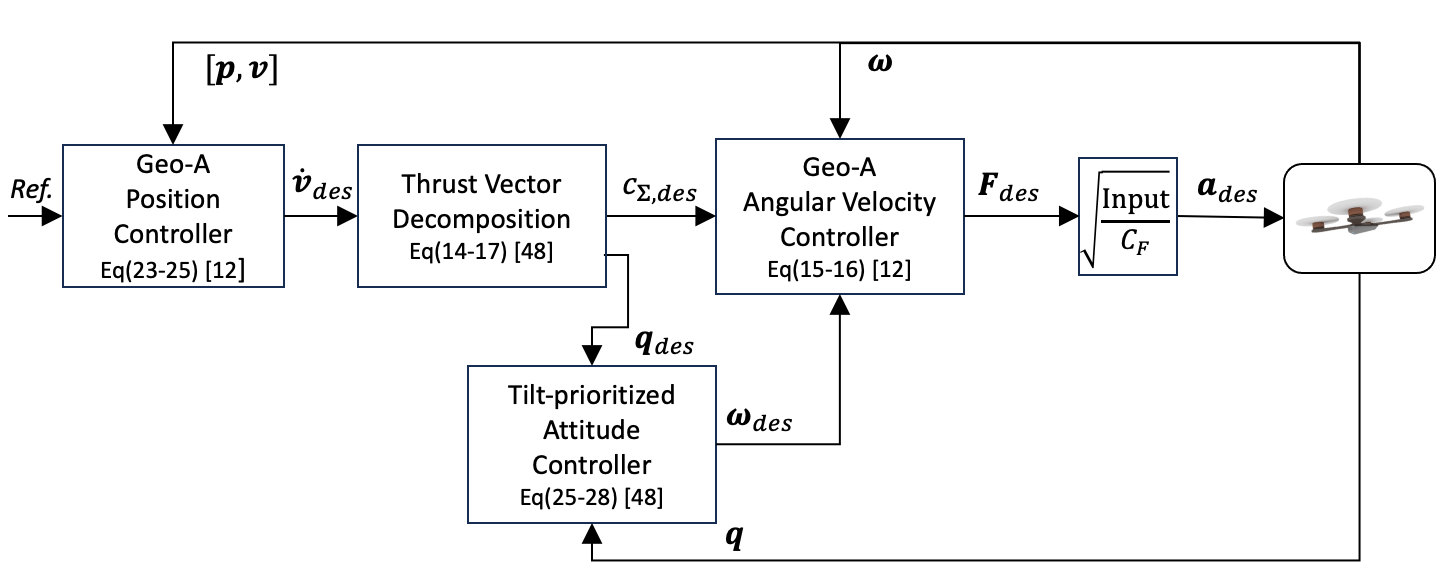}
    \caption{\rebuttal{The control diagram of \textit{Geo-A}.}}
    \label{fig:geo-a-control-diag}
\end{figure}
This baseline is a geometric adaptive controller both on the high level and low level, abbreviated as \textit{Geo-A}, an exception to our naming convention. We choose \textit{Geo-A} as a baseline to contextualize our proposed approach. This method ensures precise trajectory tracking and robust stabilization for quadrotors by adapting to uncertainties and disturbances in a geometrically consistent manner. 

We implement the geometric adaptive controller based on~\cite{goodarzi2015geometric}, replacing the original geometric attitude controller with a tilt-prioritized control from~\cite{sun2024comparativestudynonlinearmpc} to enable tracking of dynamically infeasible trajectories. Figure~\ref{fig:geo-a-control-diag} provides an overview. 
The position controller, based on (23-25) in~\cite{goodarzi2015geometric}, computes target acceleration, which is decomposed into target thrust and target attitude using (14-17) in~\cite{sun2024comparativestudynonlinearmpc}. The attitude is processed by tilt-prioritized control ((25-28) in~\cite{sun2024comparativestudynonlinearmpc}), and the resulting angular velocity and thrust are sent to the angular velocity controller ((15-16) in~\cite{goodarzi2015geometric}). Finally, thrust commands are mapped to motor speeds.

\subsection{PID-INDI-A}
\begin{figure}
    \centering
    \includegraphics[width=\linewidth]{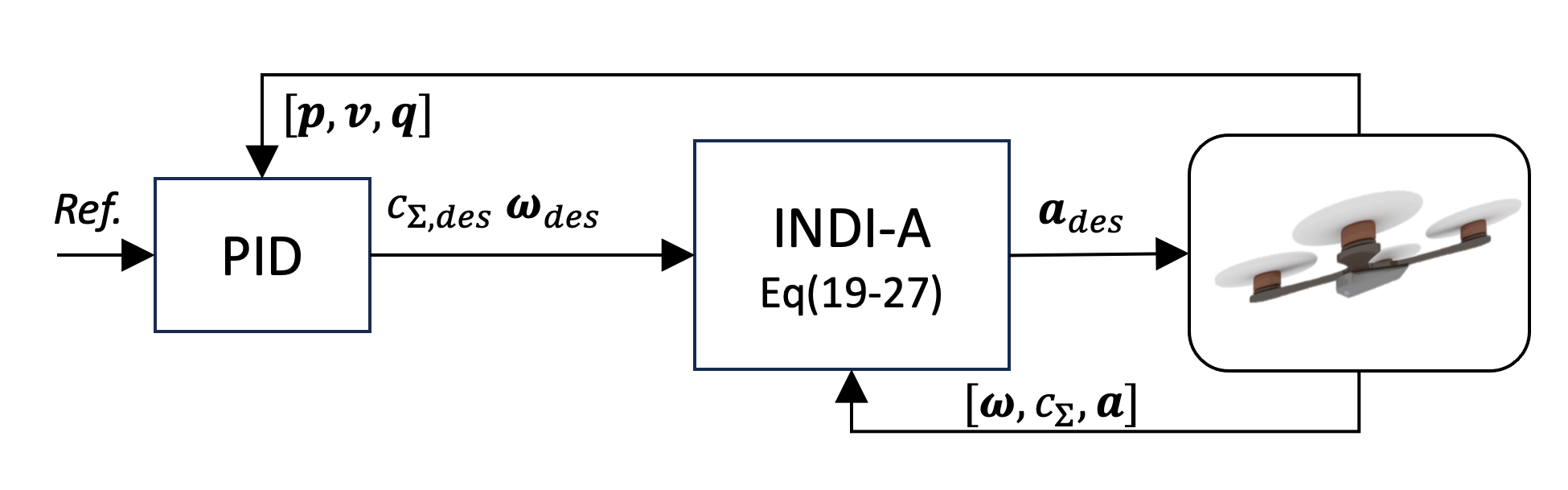}
    \caption{\rebuttal{The control diagram of \textit{PID-INDI-A}.}}
    \label{fig:indi-control-diag}
\end{figure}
\subsubsection{INDI-A}
        
We implement the adaptive INDI controller based on~\cite{smeur2015adaptINDI} and the block diagram is detailed in Figure~\ref{fig:indi-control-diag}. INDI is a sensor-based control which uses instant sensor measurement to represent the system dynamics so that to be more robust to unmodelled uncertainties in the rotation dynamics. To enhance its adaptiveness, \cite{smeur2015adaptINDI} introduces onboard adaptive parameter estimation to update control effectiveness. 

The original implementation requires measurements of angular acceleration, motor speed, and motor speed acceleration. The setup provides more informative observations than other baselines and our method, which do not have access to motor feedback. For a fair comparison, we modify the control law to reduce reliance on motor feedback, using only motor speed estimates obtained through a first-order response model with a time constant of 10ms while preserving the original linear update adaptation law.

After knowing the desired mass-normalized thrust and angular velocity from the high-level controller, we can compute the desired torque using~\eqref{eq:des-torque}. Then we can compute the desired motor speed command using~\eqref{eq:thrust-mapping} and~\eqref{eq:spd-mapping}, yielding: 
\begin{align}
     \begin{bmatrix}
        m c_{\Sigma,\mathrm{des}}\\
        \bm \tau_{\mathrm{des}}
    \end{bmatrix}&= M^{-1} \bm F_{\mathrm{des}}\\
    &= M^{-1} C_F \bm a^2_{\mathrm{des}}
\end{align}
We want to use instantaneous sensor feedback to compute the desired motor speed to be against model uncertainty and disturbance. Therefore, we decompose the thrust and torque into current measurement and desired value (see (30) and (31) of~\cite{tal2021INDI} for detailed derivations of torque decomposition). 
\begin{align}
c_{\Sigma,\mathrm{des}} &= c_{\Sigma} + (c_{\Sigma,\mathrm{des}}-c_{\Sigma})\\
    \bm \tau_{\mathrm{des}} &= \bm{\tau} + J(\bm{\dot{\omega}}_\mathrm{des} - \dot{\bm \omega}) 
\end{align}
So that the effect of unmodeled rotational dynamics is captured by the measurement of angular acceleration $\dot{\bm{\omega}}$ and torque $\bm \tau$. Note the desired angular acceleration is gained through~\eqref{eq:des-omega-dot}.

Similarly, the desired motor speed is decomposed as: 
\begin{align}
    \begin{bmatrix}
        m c_{\Sigma,\mathrm{des}}\\
        \bm \tau_{\mathrm{des}}
    \end{bmatrix}
    &=  M^{-1} C_F \bm a^2 +M^{-1} C_F (\bm a^2_{\mathrm{des}} - \bm a^2)
\end{align}
By rearranging the terms, we obtain the following equation to solve for $\bm a_{\mathrm{des}}$:
\begin{align} \label{eq:indi-a-unsimp}
\bm a_\mathrm{des}^2 - \bm a^2 &=C_F^{-1}M \begin{bmatrix}
     m (c_{\Sigma,\mathrm{des}}-c_{\Sigma})\\
       J (\bm{\dot{\omega}}_\mathrm{des} - \dot{\bm \omega})
    \end{bmatrix} 
\end{align}
This equation can then be reformulated into the INDI framework by introducing the control allocation matrices $\bm G_1$ and $\bm G_2$: 
\begin{align}
    \bm a_\mathrm{des}^2 - \bm a^2 &= \bm G_1 (c_{\Sigma,\mathrm{des}} - c_{\Sigma}) + \bm G_2 (\bm{\dot{\omega}}_\mathrm{des} - \bm{\dot{\omega}})
\end{align}
with the adaptation law defined as
\begin{align}
    \bm G(k) &= \bm G(k-1) - \mu \bm G(k-1)  \left(\begin{bmatrix}
        \Delta c_{\Sigma} \\ \Delta \bm{\dot{\omega}}
    \end{bmatrix} - \Delta \bm (a^2) \right) \begin{bmatrix}
        \Delta c_{\Sigma} \\ \Delta \bm{\dot{\omega}}
    \end{bmatrix}^T \\
    \bm G &= \begin{bmatrix}
        \bm G_1 &\bm G_2
    \end{bmatrix}
\end{align}
where $\Delta$ denotes the difference between the current and previous sampled variables. A more detailed analysis on the stability and performance of this method could be found in~\cite{smeur2015adaptINDI}.



\bibliographystyle{IEEEtran}
\bibliography{references}
\begin{IEEEbiography}[{\includegraphics[width=1in,height=1.25in,clip,keepaspectratio]{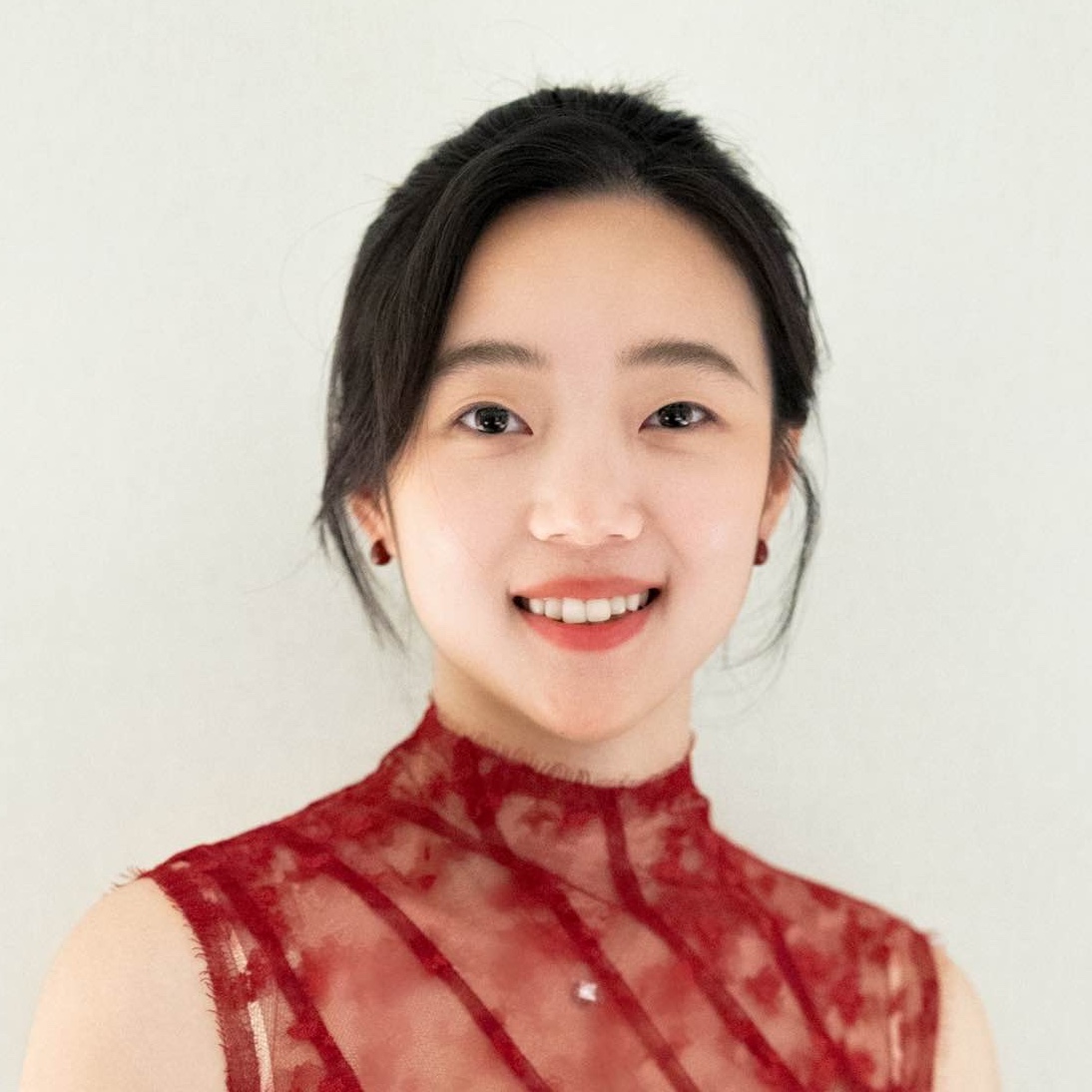}}]{Dingqi Zhang} is currently a PhD student in the High Performance Robotics Laboratory at UC Berkeley. Her research focuses on extreme adaptive control for aerial vehicles with data-driven methods. Her work spans the intersection of dynamics, control, and system design, with an emphasis on building robust controllers that remain reliable across hardware changes, disturbances, and structural variations. She previously received her B.S. in Mechanical Engineering and Computer Science from Cornell University in 2021.
\end{IEEEbiography}
\begin{IEEEbiography}
[{\includegraphics[width=1in,height=1.25in,clip,keepaspectratio]{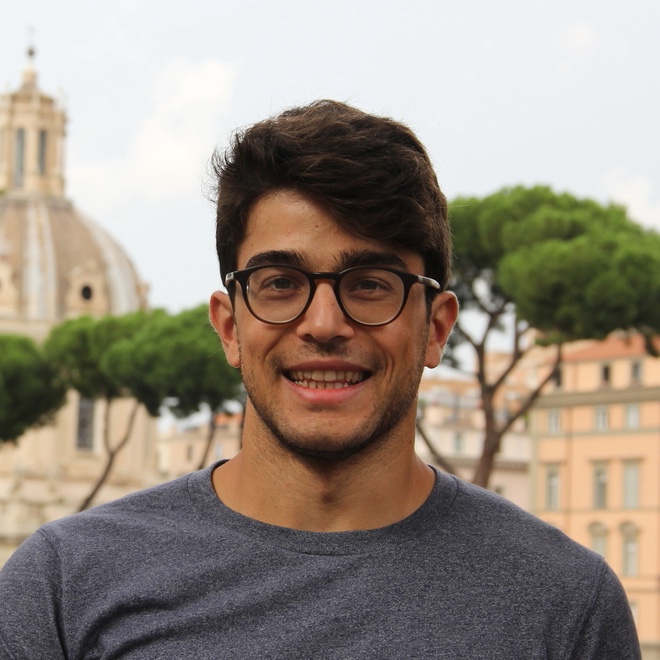}}]{Antonio Loquercio} is a professor of electrical engineering and computer science at the University of Pennsylvania. He received a M.Sc. degree from ETH Zurich and a Ph.D. from the University of Zurich in 2017 and 2021, respectively. He worked at the Berkeley Artificial Intelligence Research Lab (BAIR) at UC Berkeley from 2022 to 2024.
\end{IEEEbiography}
\begin{IEEEbiography}
[{\includegraphics[width=1in,height=1.25in,clip,keepaspectratio]{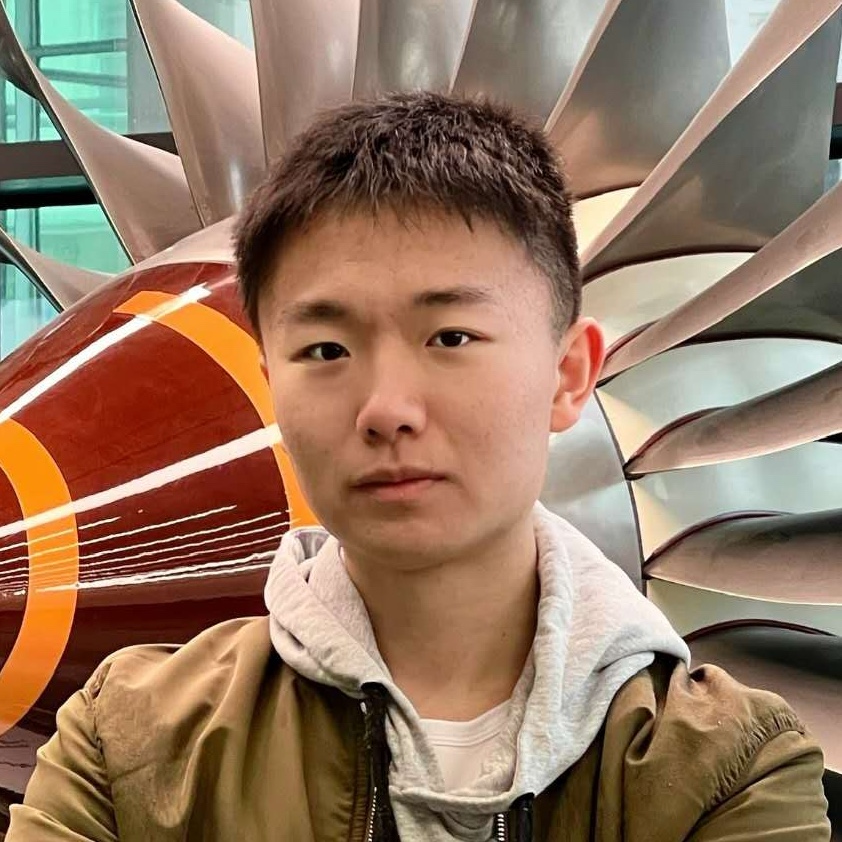}}]{Jerry Tang} graduated from UC
Berkeley with a Ph.D. in Mechanical Engineering in 2025. He received his Bachelor of Science degree in Mechanical Engineering from UC Berkeley in 2020, and his Master of Science degree in Mechanical Engineering from UC Berkeley in 2021. His research interest includes the design and control of novel aerial robots.
\end{IEEEbiography}
\begin{IEEEbiography}
[{\includegraphics[width=1in,height=1.25in,clip,keepaspectratio]{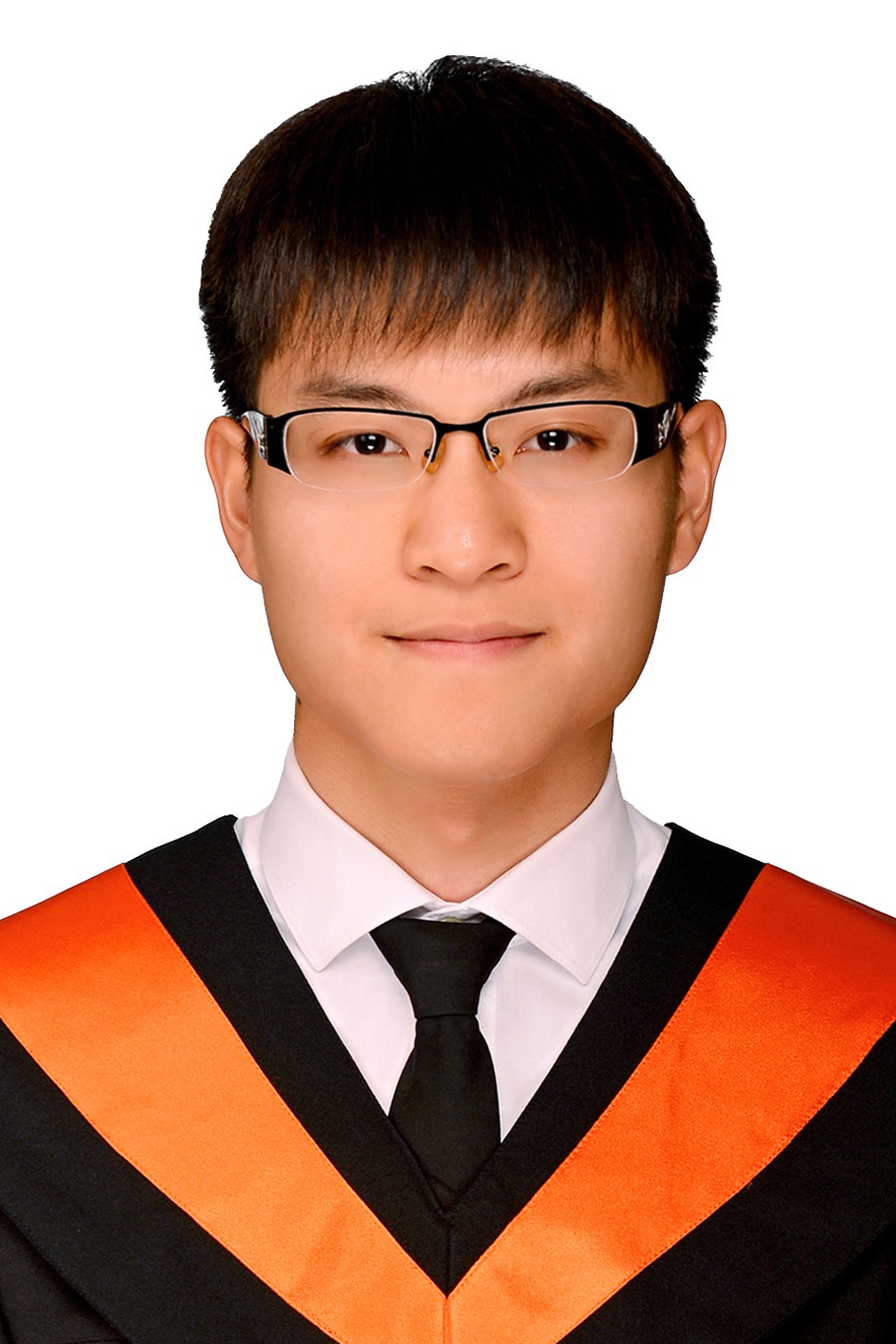}}]{Ting-Hao Wang} received his B.S. and M.S. degrees in Mechanical Engineering from National Taiwan University in 2016 and 2018, respectively. He is currently pursuing a Ph.D. in Mechanical Engineering at the University of California, Berkeley. His research focuses on efficient path planning for aerial systems operating in complex and cluttered environments, with an emphasis on platforms constrained by limited sensing and computational capabilities. He is particularly interested in streamlining autonomy systems to address key challenges in real-time decision-making and safe navigation for aerial robots deployed in resource-limited and dynamically changing environments.  
\end{IEEEbiography}
\begin{IEEEbiography}
[{\includegraphics[width=1in,height=1.25in,clip,keepaspectratio]{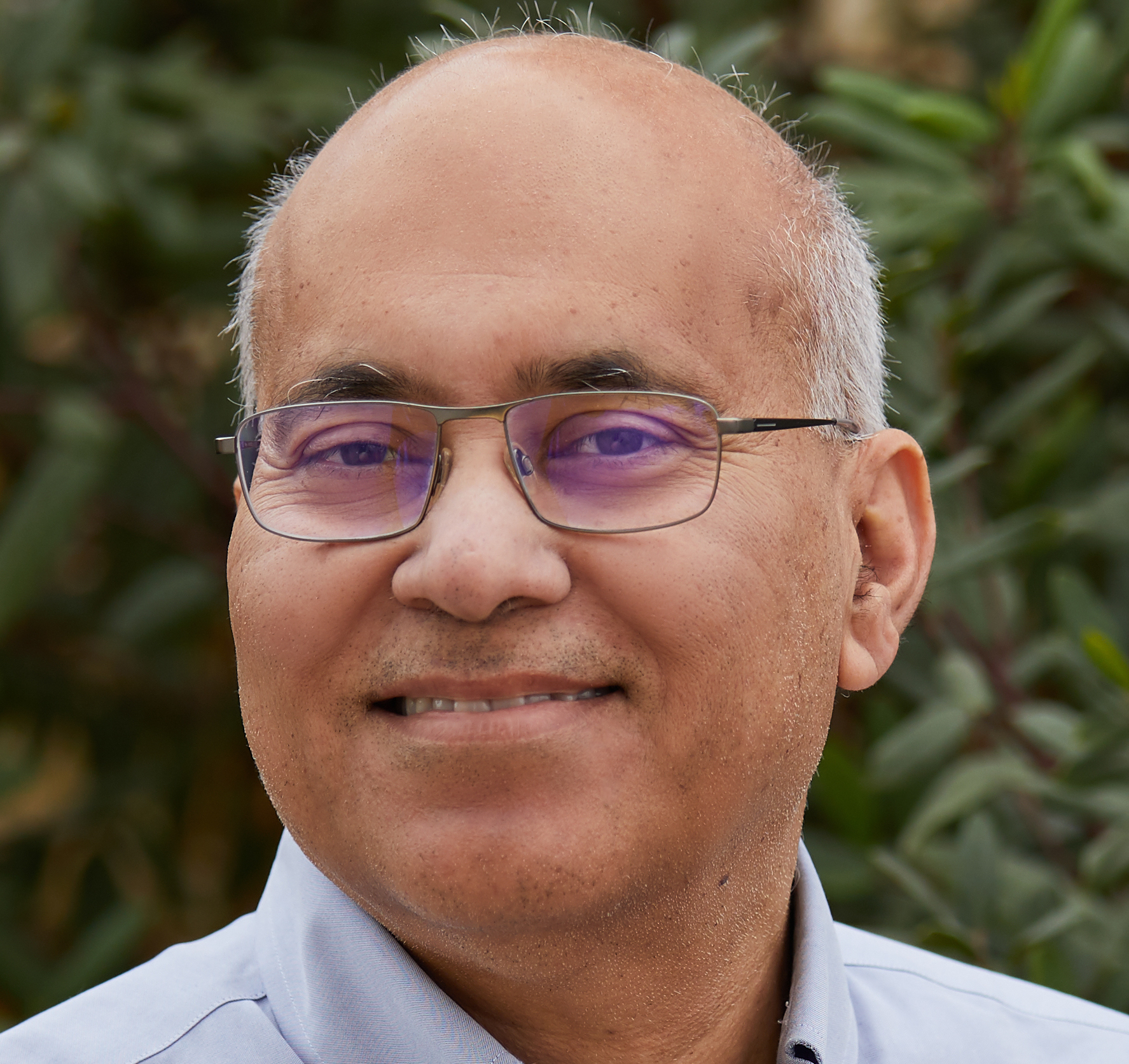}}]{Jitendra Malik} is Arthur J. Chick Professor of EECS at UC Berkeley and Vice-President, Robotics Research at Meta, Inc. His group has worked on many topics in computer vision, robotics and machine learning, receiving eleven best paper awards, including six test of time awards.  He has mentored more than 80 PhD students and postdoctoral fellows, who have gone on to become leading researchers at places like MIT, Berkeley, CMU, Caltech, Cornell, UIUC, UPenn, Michigan, UT Austin, Google and Meta. Jitendra received the 2013 IEEE PAMI-TC Distinguished Researcher in Computer Vision Award, the 2016 ACM-AAAI Allen Newell Award, the 2018 IJCAI Award for Research Excellence in AI, and the 2019 IEEE Computer Society Computer Pioneer Award. He is a fellow of IEEE and ACM. He is a member of the National Academy of Engineering and the National Academy of Sciences, and a fellow of the American Academy of Arts and Sciences.
\end{IEEEbiography}
\begin{IEEEbiography}
[{\includegraphics[width=1in,height=1.25in,clip,keepaspectratio]{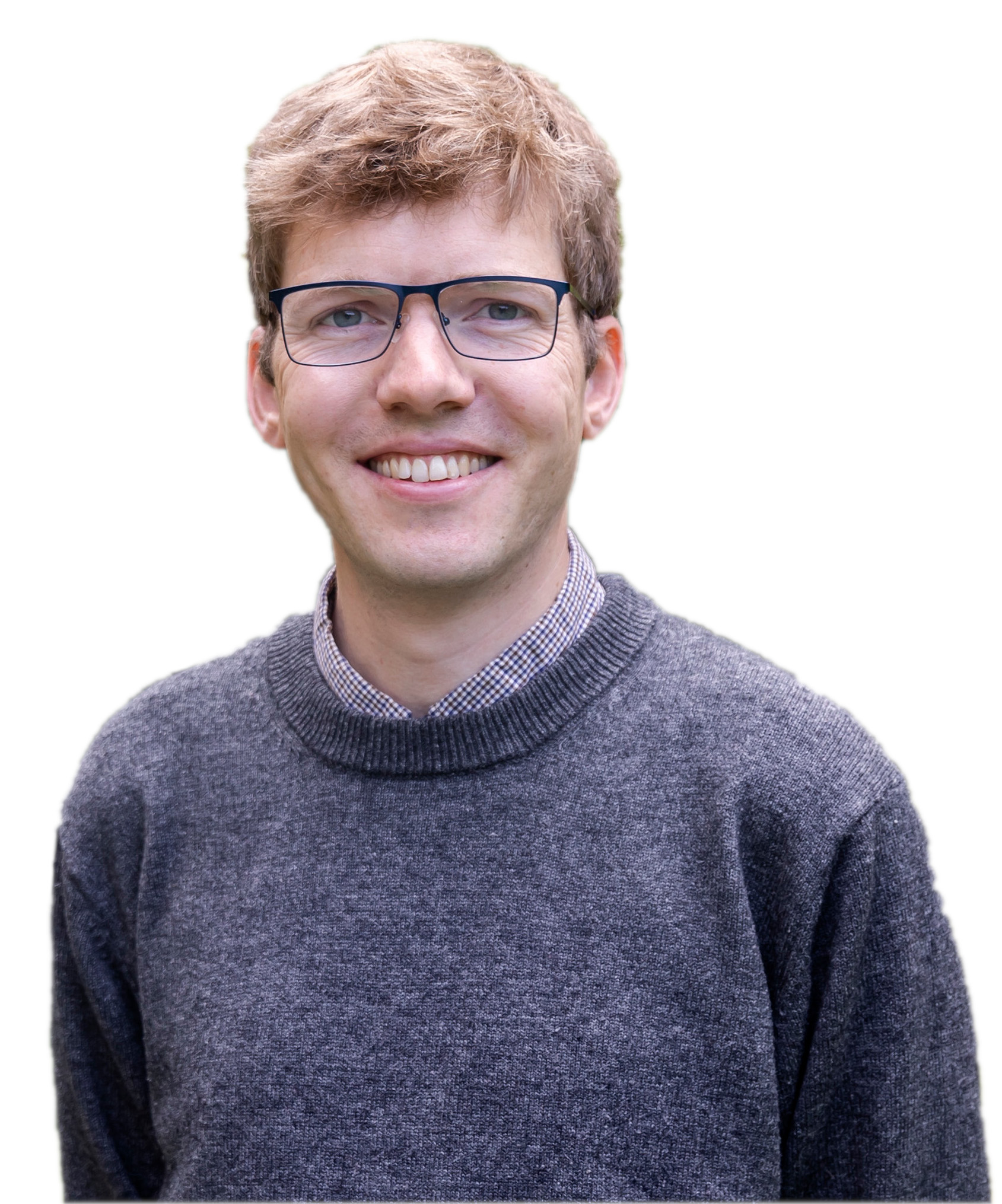}}]{Mark W. Mueller} is an associate professor of Mechanical Engineering at the University of California, Berkeley, and runs the High Performance Robotics Laboratory (HiPeRLab). He received a Dr.Sc. and M.Sc. from the ETH Zurich in 2015 and 2011, respectively, and a BSc from the University of Pretoria in 2008. His research interests include aerial robotics, their design and control, and especially the interactions between physical design and algorithms.
\end{IEEEbiography}

\end{document}